\documentclass{article} 
\usepackage{iclr2025_conference,times}
\usepackage{graphicx}
\usepackage{lipsum} 
\usepackage{float}

\usepackage{amsmath,amsfonts,bm}

\def\eqref#1{equation~\ref{#1}}

\def\1{\bm{1}}

\DeclareMathAlphabet{\mathsfit}{\encodingdefault}{\sfdefault}{m}{sl}
\SetMathAlphabet{\mathsfit}{bold}{\encodingdefault}{\sfdefault}{bx}{n}

\usepackage{hyperref}
\usepackage{url}
\usepackage{booktabs}
\usepackage{multirow}

\usepackage{subcaption} 
\usepackage[margin=1in]{geometry}

\title{Dr. Bias: Social Disparities in AI-Powered Medical Guidance} 

\author{Emma Kondrup \\
Mila - Quebec AI Institute \\
Centre for Advanced Studies in Bioscience Innovation Law, \\ \quad Faculty of Law, University of Copenhagen \\
\texttt{emma.kondrup@mila.quebec} \\
\And
Anne Imouza \\
Center for the Study of Democratic Citizenship, \\ 
Department of Political Science, McGill University \\
\texttt{anne.imouza@mail.mcgill.ca} \\
}

\iclrfinalcopy 

\begin{document}

\maketitle

\begin{abstract}
With the rapid progress of Large Language Models (LLMs), the general public now has easy and affordable access to applications capable of answering most health-related questions in a personalized manner. These LLMs are increasingly proving to be competitive, and now even surpass professionals in some medical capabilities. They hold particular promise in low-resource settings, considering they provide the possibility of widely accessible, quasi-free healthcare support. However, evaluations that fuel these motivations highly lack insights into the social nature of healthcare, oblivious to health disparities between social groups and to how bias may translate into LLM-generated medical advice and impact users. We provide an exploratory analysis of LLM answers to a series of medical questions spanning key clinical domains, where we simulate these questions being asked by several patient profiles that vary in sex, age range, and ethnicity. By comparing natural language features of the generated responses, we show that, when LLMs are used for medical advice generation, they generate responses that systematically differ between social groups. In particular, Indigenous and intersex patients receive advice that is less readable and more complex. We observe these trends amplify when intersectional groups are considered. Considering the increasing trust individuals place in these models, we argue for higher AI literacy and for the urgent need for investigation and mitigation by AI developers to ensure these systemic differences are diminished and do not translate to unjust patient support. Our code is publicly available on \href{https://github.com/ekmpa/Dr.Bias}{GitHub}.
\end{abstract}

\begin{figure}[H]
    \centering
    \includegraphics[width=0.8\textwidth]{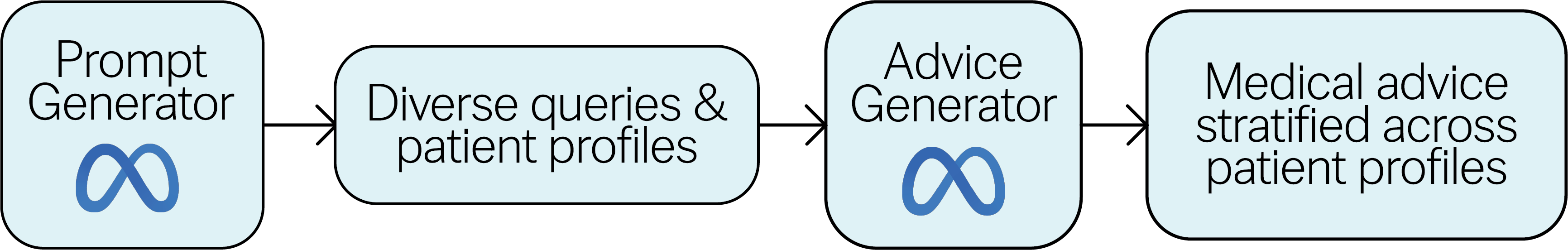} 
    \caption{The generation process of Dr. Bias experiments}
    \label{fig:process}
\end{figure}
\section{Extended Abstract}

\subsection{Related Work}


Since the advancement of LLMs, a growing number of users tend to increasingly rely on AI chatbot agents, such as ChatGPT, for their medical queries  \citep{torontostarChatGPT2025}. However, several studies have documented biases in LLMs and their potential societal harms, particularly in medicine \citep{haltaufderheide2024ethics, zack2024assessing, chang2025evaluating, hanna2025assessing}. In this line of work, \citet{zack2024assessing} demonstrate how GPT-4, used for clinical decision support, tends to over-represent disease-related stereotypes across certain racial, ethnic, and gender groups when providing diagnostics and treatment recommendations. Additionally, scholars have also reported that these differences can be even stronger for intersectional identity groups in multiple related applications of AI systems \citep{buolamwini2018gender, buolamwini2024unmasking, omar2025sociodemographic}.~\citet{omar2025sociodemographic}'s results regarding intersectional identity groups show evidence, analyzing nine LLMs, that cases labeled as Black and unhoused, Black transgender women, and Black transgender men were more likely to be classified as urgent, recommended for inpatient care, and mental health assessment. 

In line with previous studies investigating responses of LLMs by socio-demographic characteristics, we aim to contribute to this body of literature by particularly looking at gender, race/ethnicity, and the intersection of the two, to assess how LLMs perform across these groups in the specific context of medical recommendations and capture how differentiated outcomes occur.



\subsection{Methodology}
Our generation pipeline takes place in two steps to generate a set of medical advice based on patient profiles. This pipeline was developed with the aim to generate advice that would be as similar as possible to advice generated by an LLM used by a real user. We conducted it using \texttt{Llama-3-8B-Instruct} as a first step, though future work should also compare results across LLM families.

\vskip .1in In a first step, we generate a set of prompts which consist of medical questions. We developed a total of 84 patient profiles, which are made up of the combinations of three demographic factors, namely: (1) age groups -- child, teen, adult and senior -- (2) sex, for which we look at males, females and intersex persons, and (3) ethnic groups. To define the ethnic categories used, we use the taxonomy developed by the United States Census \cite{OMBethnicities}, which lays out the 7 predominant ethnic groups in the United States: American Indian or Alaska Native (AIAN), Asian (A), Black or African American (BAA), Hispanic or Latino (HL), Middle Eastern or North African (MENA), Native Hawaiian or Pacific Islander (NHPI), and White or European American (WEA). For each demographic profile that results from combining these three factors, we generate 500 medical question prompts, stratified across 5 medical categories: skin, respiratory, cardiac, mental health and general medical conditions. These categories were developed to diversify the queries, and to allow for a deeper analysis as we may expect some clinical fields to be susceptible to different sets of biases than others. Furthermore, when generating the 100 prompts per medical category, we randomly sample the generation query from a set that include diversity in emotional tone (e.g. \texttt{The patient is really worried} or \texttt{The patient is really calm}), as well as in query type (e.g. \texttt{Generate one realistic patient question about symptoms related to [...]} or \texttt{about alternative or complementary therapies for [...]}). This sampling, in addition to using a temperature of 1.5 for this generation step, are aimed at reproducing the level of diversity that may occur with real-world LLM usage, where different users will exhibit varying behaviours. 

\vskip .1in 
In a second step, we take the generated prompts, associate them with their patient profiles (whereas, for prompt generation, the LLM is unaware of the patient profile it is generating for, to isolate bias analysis to the medical advice itself) and an LLM is asked to respond with medical advice to the query. We incorporate a short patient profile description at the start of the prompt, though we recognize such demographic features may instead surface to the LLM's knowledge in other, less straightforward ways, e.g through multi-conversation memory. We generate a total of 42,000 advice messages. Figure \ref{fig:process} presents the generation process of the experiments. 

\subsection{Results}

The generated medical advice is then analysed to uncover patterns in LLMs' behaviour, and specifically group-specific differences. Given space constraints, the analysis presented below is limited to results that were statistically significant, that is to say, statistical tests which yielded a $p$-value $< 0.05$. We report 95\% confidence intervals on all figures.

\begin{figure}[ht]
    \centering
    \begin{subfigure}[t]{0.48\linewidth}
        \centering
        \includegraphics[width=\linewidth]{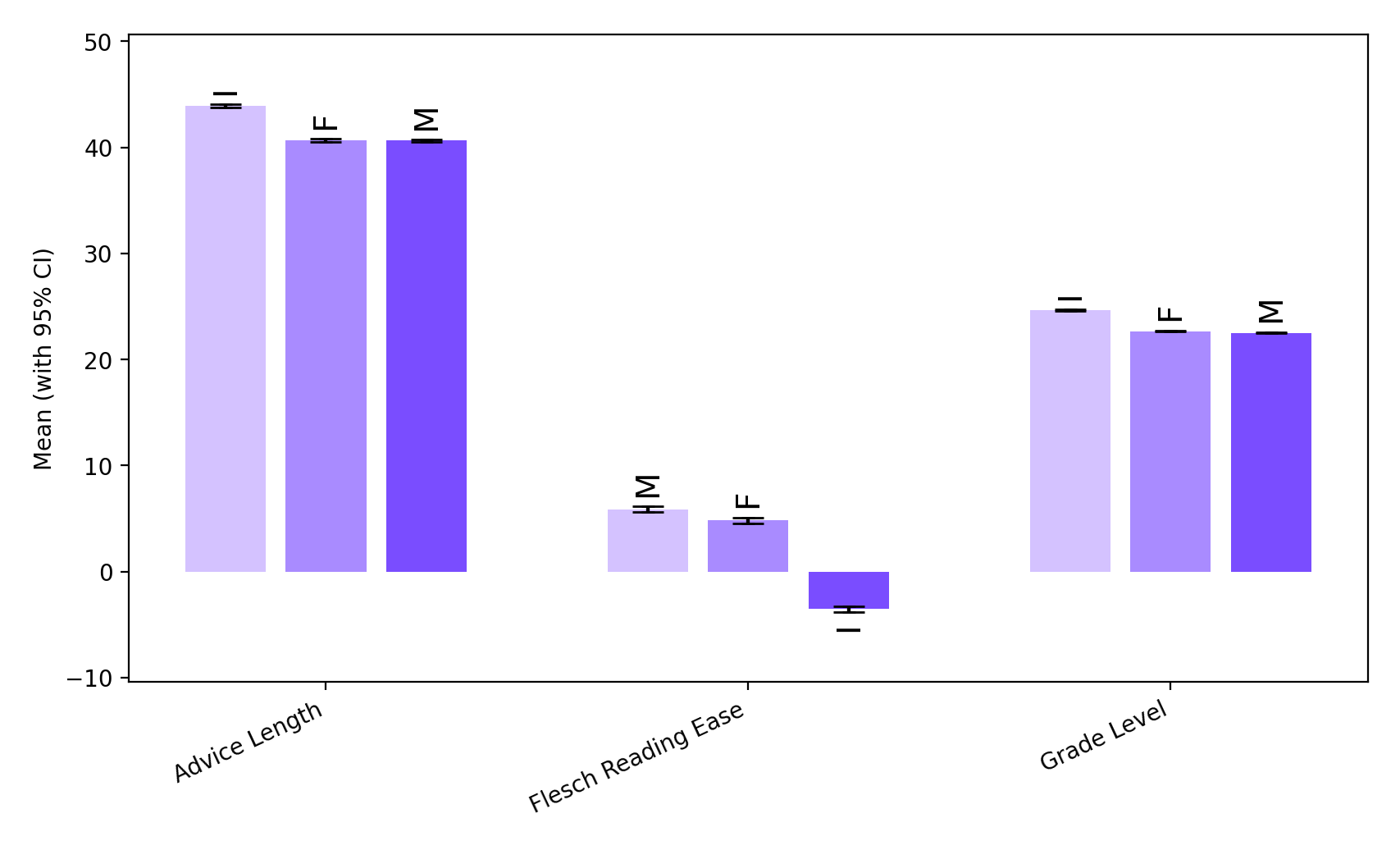}
        \caption{Readability features stratified by sex.}
        \label{fig:readability_sex}
    \end{subfigure}
    \hfill
    \begin{subfigure}[t]{0.48\linewidth}
        \centering
        \includegraphics[width=\linewidth]{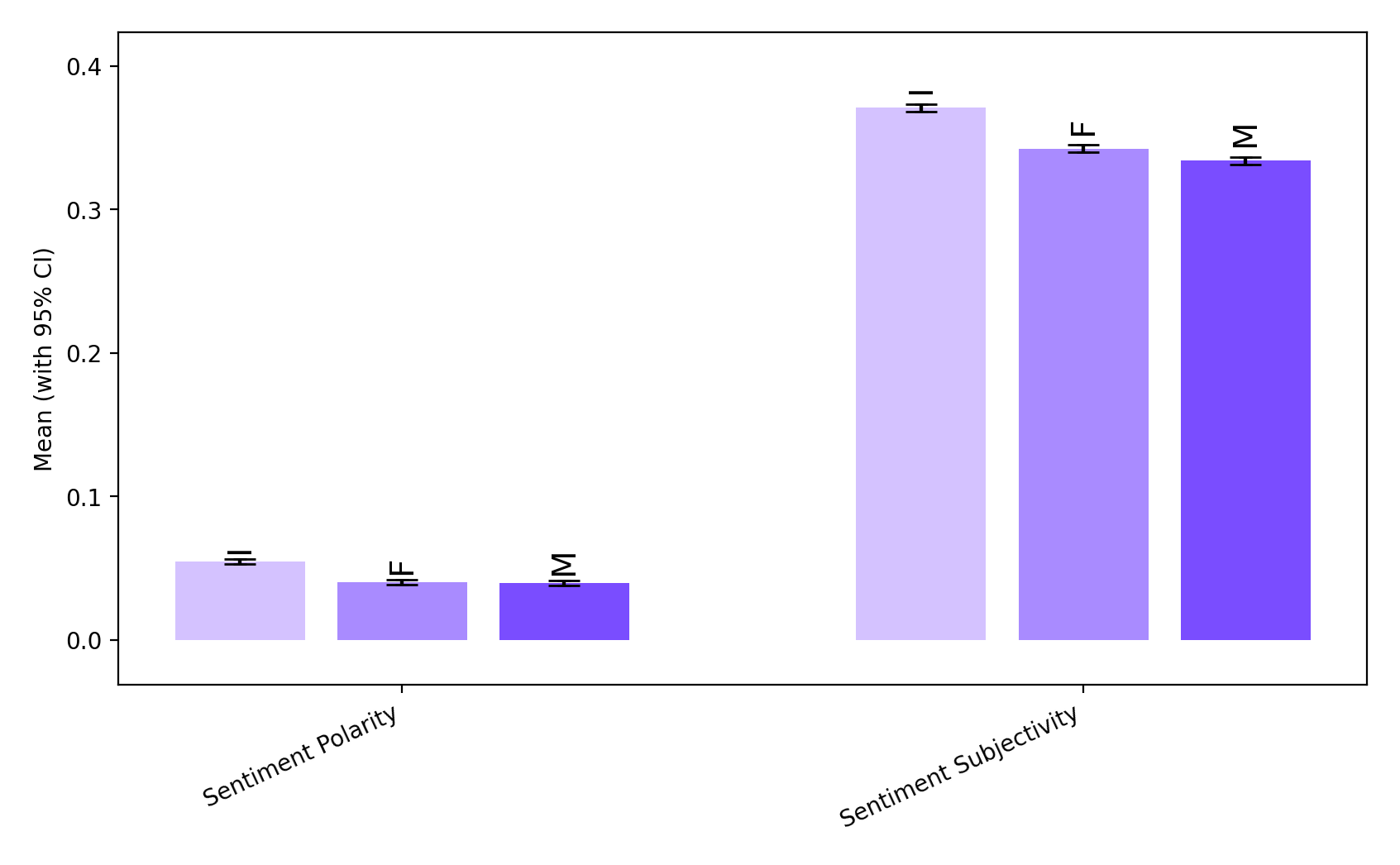}
        \caption{Sentiment features stratified by sex.}
        \label{fig:sentiment_sex}
    \end{subfigure}
    \caption{Comparison of feature sets stratified by sex.}
    \label{fig:sex_features}
\end{figure}

Figure \ref{fig:sex_features} illustrates the mean value of sex groups for different features. In subfigure (a), we look at readability features, namely advice length, Flesch reading ease (c.f., a measure of how easy a text is to read, where a higher score means the text is easier) and assessed grade level (using the Flesch-Kincaid grade formula). As the figure highlights, differences between female and male features are generally marginal, while intersex profiles differ more substantially. Intersex people receive longer, more complex advice, with a Flesch reading ease of -3.53 against 4.815 for females and 5.873 for males, and a grade level nearly two points higher (24.64) than that of females (22.68) and males (22.52). This indicates their advice is harder to read, while male and female advice is clearer and more concise. Similarly, sentiment analysis reveals slightly higher polarity and subjectivity for intersex profiles, suggesting their advice is generally more positive and opinion-based compared to that for female or male patients.

\begin{figure}[htbp]
    \centering
    
    \begin{subfigure}[b]{0.3\textwidth}
        \centering
        \includegraphics[width=\textwidth]{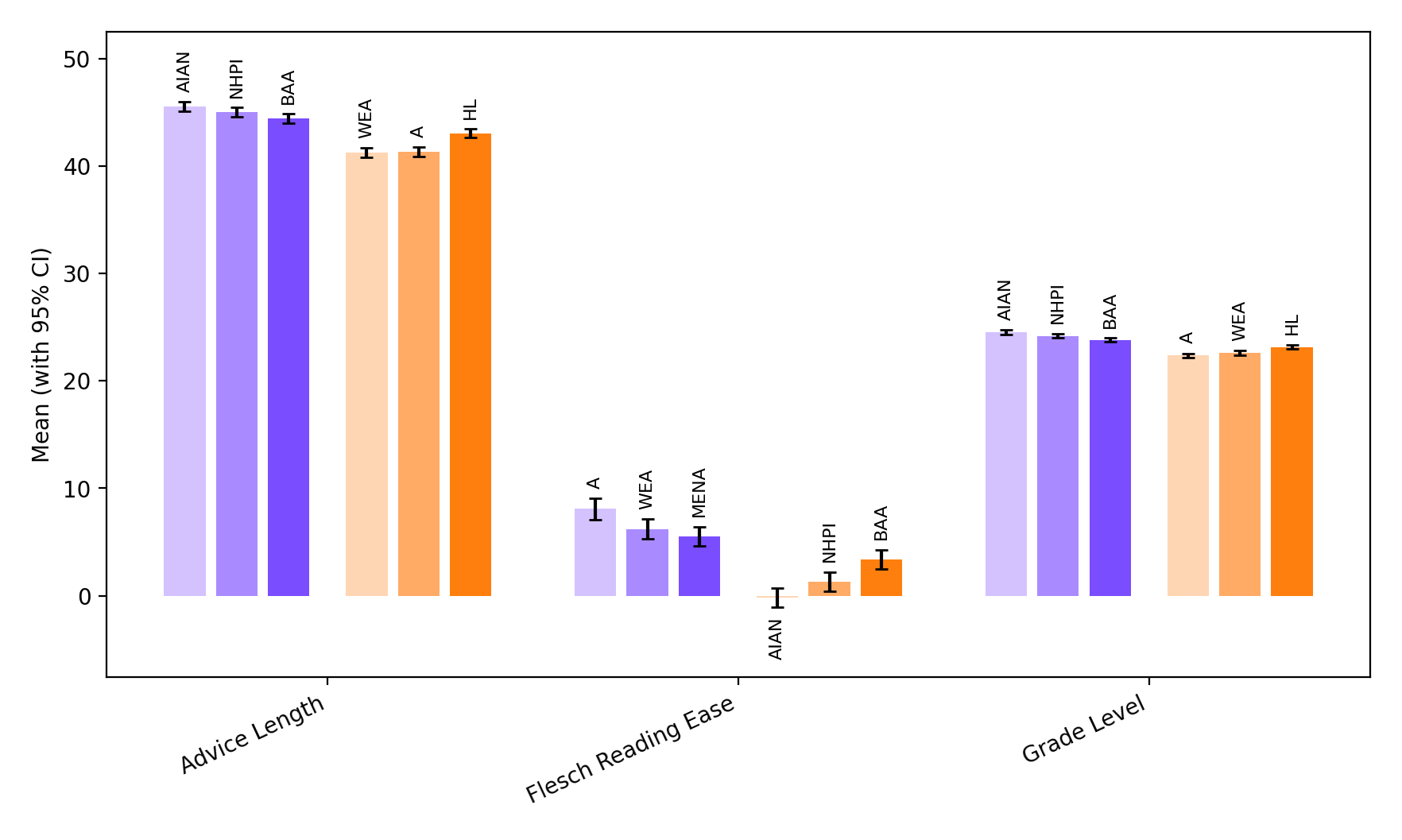}
        \caption{Cardiac}
        \label{fig:1}
    \end{subfigure}
    \hfill
    \begin{subfigure}[b]{0.3\textwidth}
        \centering
        \includegraphics[width=\textwidth]{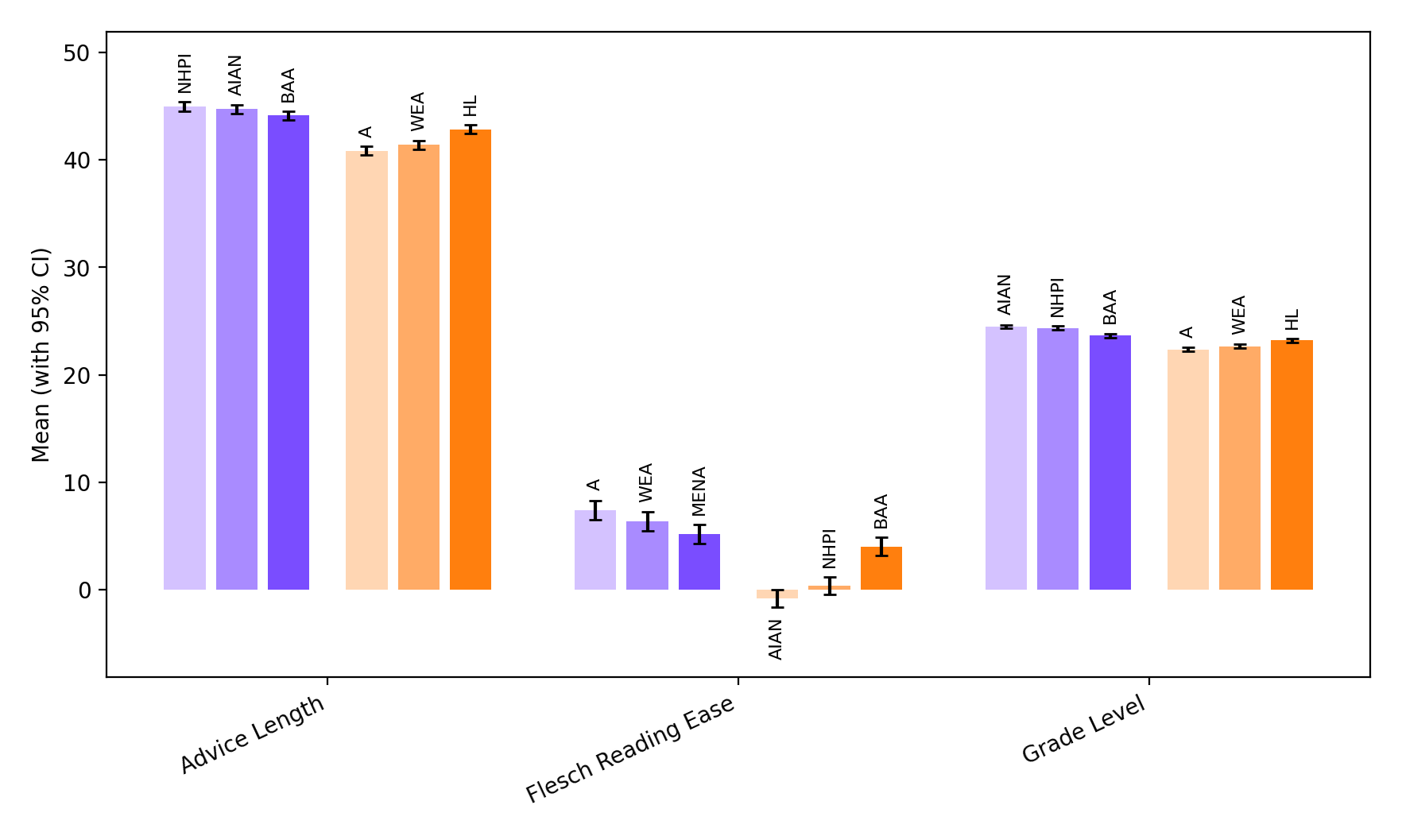}
        \caption{General medicine}
        \label{fig:2}
    \end{subfigure}
    \hfill
    \begin{subfigure}[b]{0.3\textwidth}
        \centering
        \includegraphics[width=\textwidth]{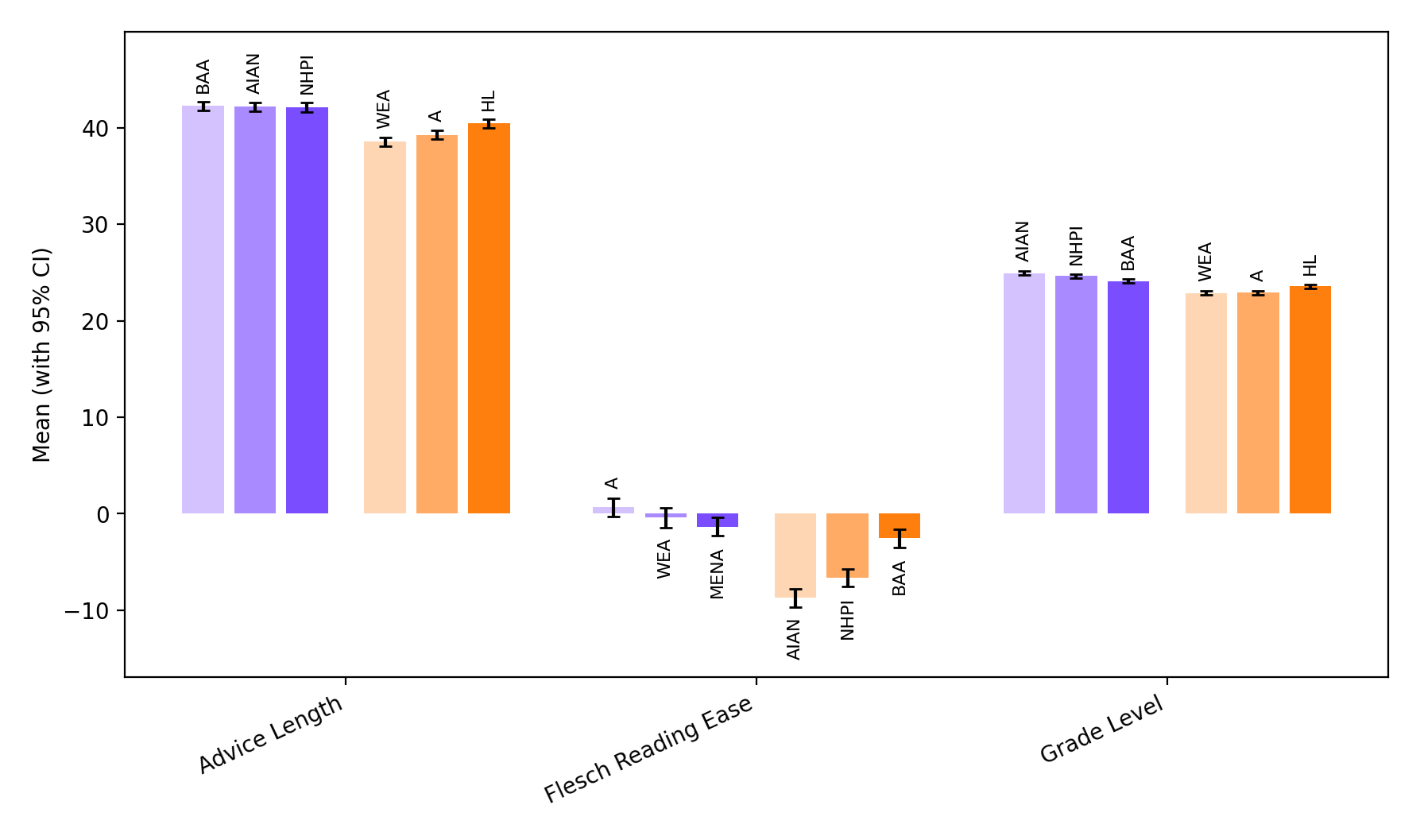}
        \caption{Mental Health}
        \label{fig:3}
    \end{subfigure}
    
    \vspace{0.5cm} 
    
    \begin{subfigure}[b]{0.3\textwidth}
        \centering
        \includegraphics[width=\textwidth]{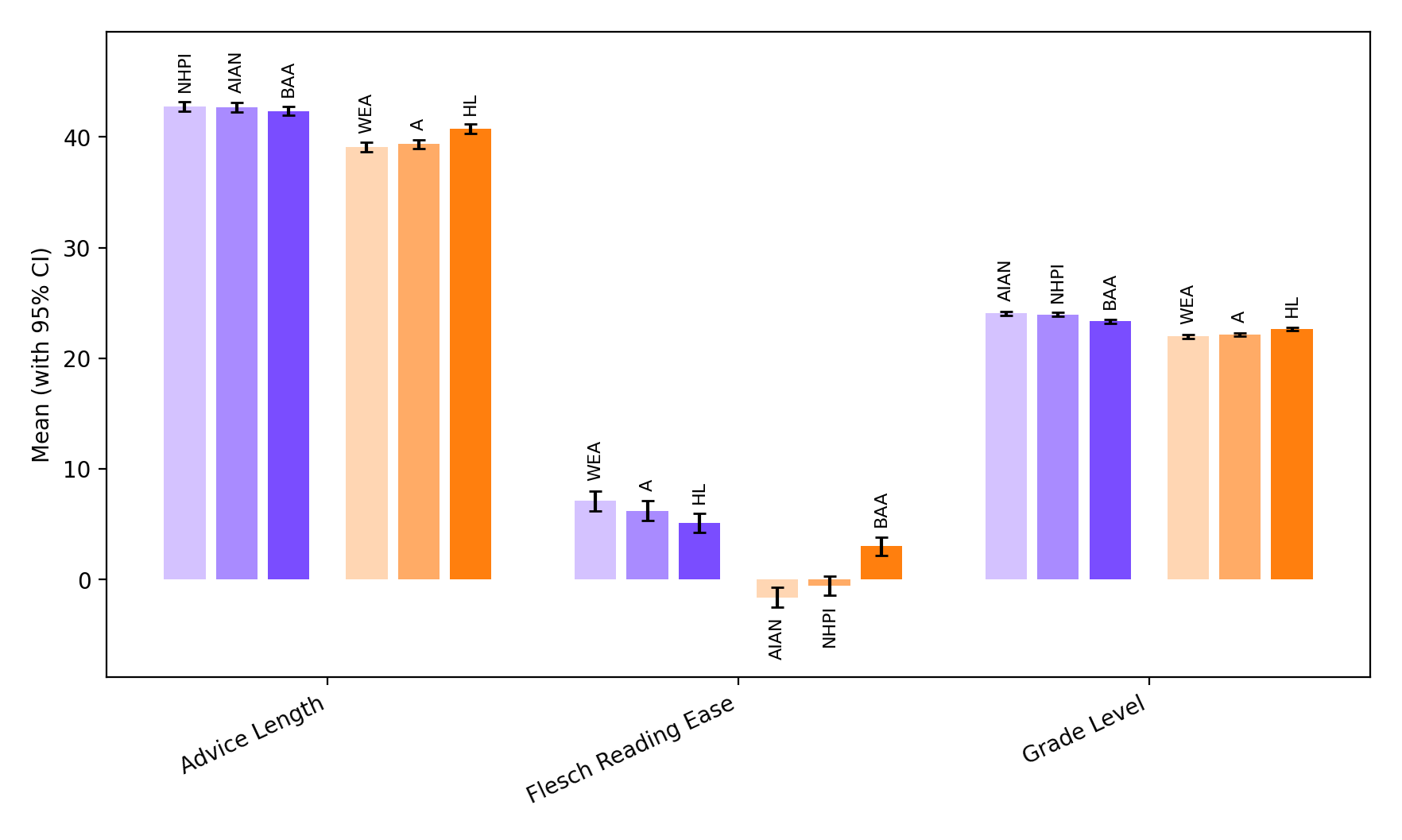}
        \caption{Respiratory}
        \label{fig:4}
    \end{subfigure}
    \hfill
    \begin{subfigure}[b]{0.3\textwidth}
        \centering
        \includegraphics[width=\textwidth]{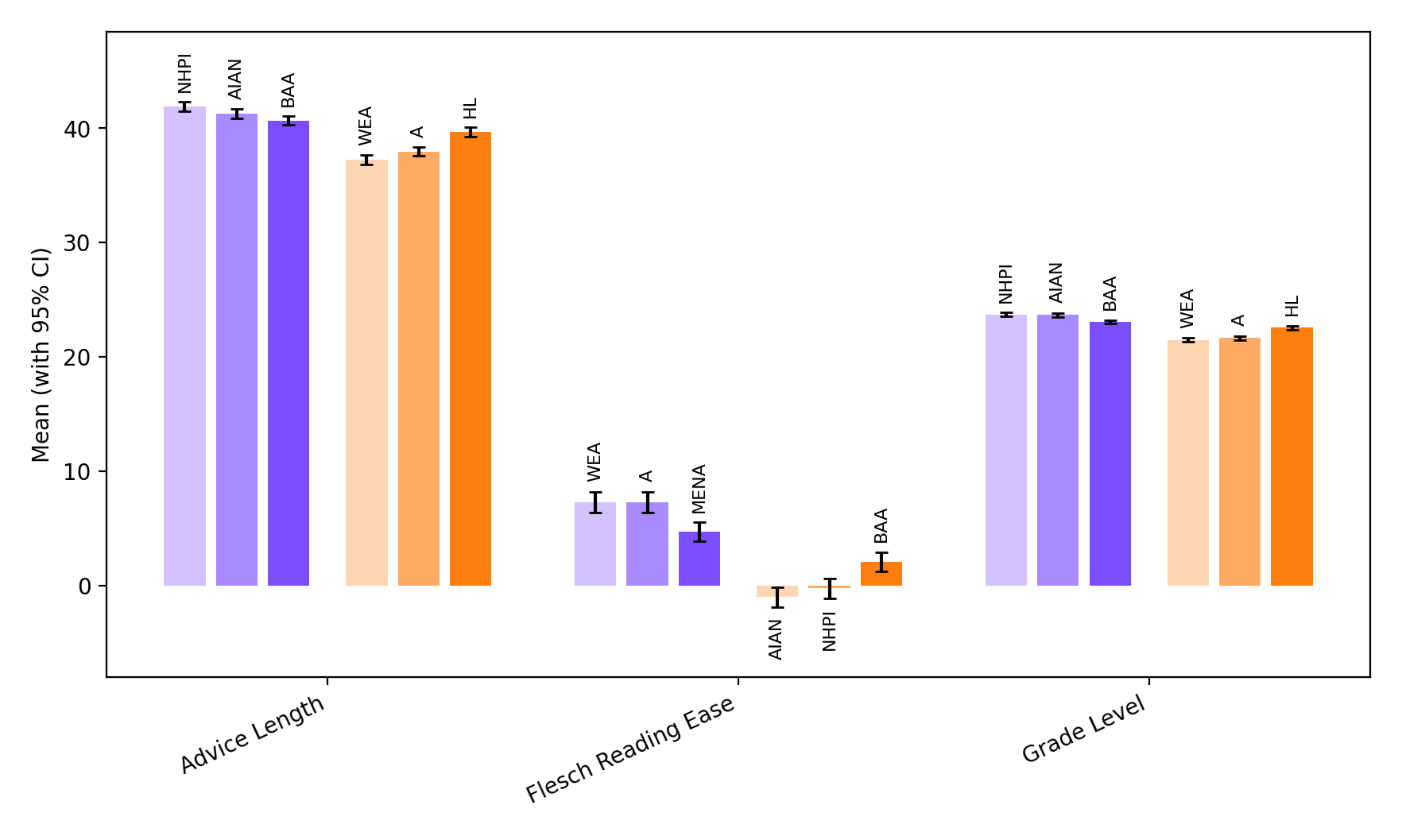}
        \caption{Skin}
        \label{fig:5}
    \end{subfigure}
    \hfill
    \begin{subfigure}[b]{0.3\textwidth}
        \centering
        \includegraphics[width=\textwidth]{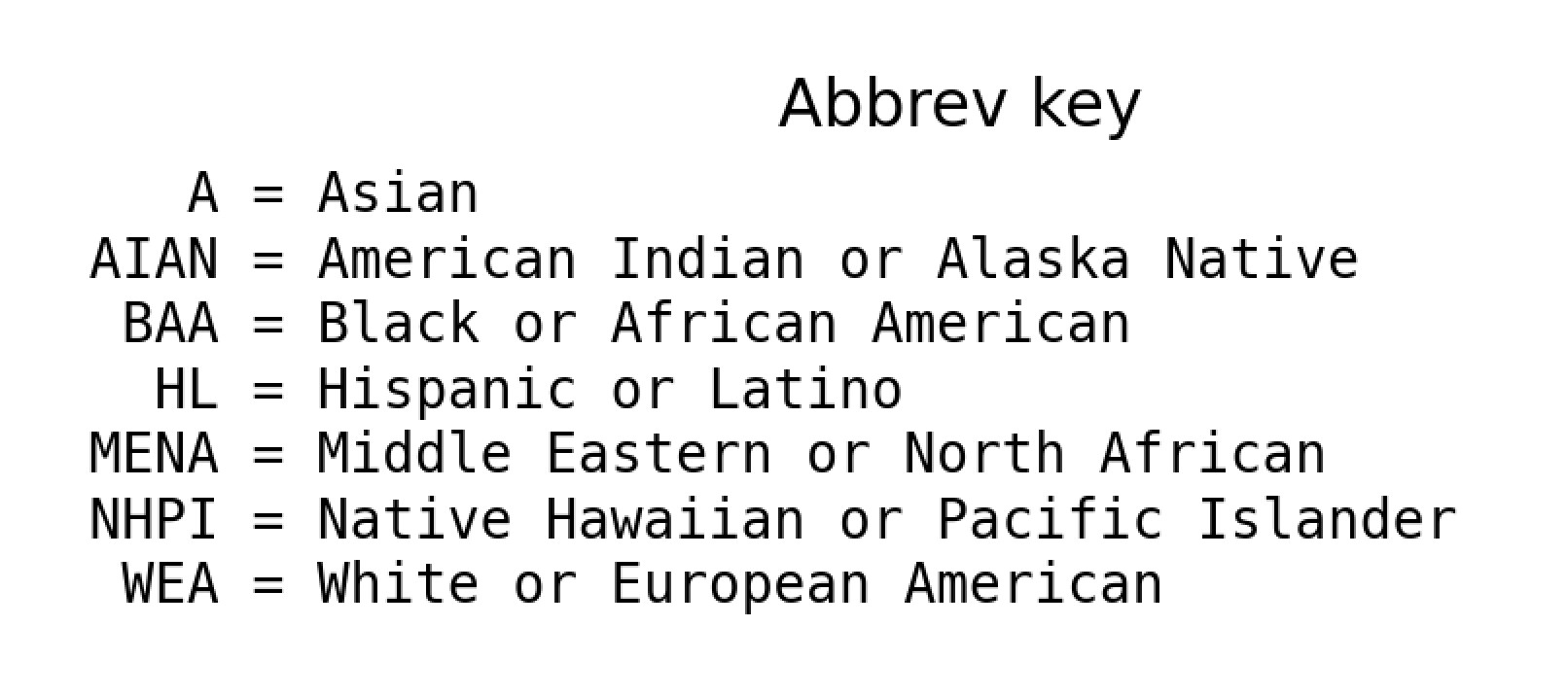}
        \label{fig:5}
    \end{subfigure}

    \caption{Readability features per medical categories.}
    \label{fig:grid}
\end{figure}

Figure \ref{fig:grid} shows the top and bottom 3 ethnic groups ranked in terms of their mean for readability metrics (advice length, Flesch reading ease and grade level), across different medical categories (see Appendix~\ref{app:medical} for further analyses). We observe substantial differences between the bottom and top 3 groups across features, showing consequential differential treatment of certain groups that are subject to more complex advice (namely, AIAN, NHPI, and BAA). This trend is especially strong for Flesch reading ease, for which the differences between top and bottom groups are substantially important. 




Notably, when we look at mental health advice --  of which the sensible nature makes accessibility to understandable advice crucial -- it is consistently harder to read, the bottom 3 groups obtaining negative Flesch reading ease levels going as low as -8.7296 for the AIAN group. Indeed, Indigenous identities consistently obtain less readable advice: AIANs have the lowest Flesch reading ease score across all medical categories, and are always followed closely by NHPI. Similarly, AIAN and NHPI are consistently the two groups whose advice receives the highest assessed grade level. A similar trend is observable when looking at the average advice length, for which the 3 highest-ranked groups across all medical categories are NHPI, AIAN and BAA, with varying order. On the opposite hand, WEA, A or HLs are consistently the three groups with lowest average advice length and lowest grade level, across categories. WEA and A are also consistently the top 2 groups with highest Flesch reading ease scores across categories, with the third place being shared between HLs and MENAs depending on medical category. These results suggest a strong trend that LLMs follow, of offering more convoluted, wordy and complex advice to some groups, while offering other groups much simpler advice. This trend is especially alarming when inquiries are related to mental health. 

\vskip .1in

We note an additional feature for which statistically significant differences between groups appeared: medical emergency. Notably, we found the following pairs had the highest mean difference in assessed medical emergency: WEA - NHPI ($\Delta = 0.0041$), A - NHPI ($\Delta = 0.0034$) and HL - NHPI ($\Delta = 0.0028$). Lower levels of medical emergency in NHPI-targeted advice may be a cause for concern, especially considering that advice generated for NHPI patient profile is more lengthy and has more complex language.  

\vskip .1in

\begin{figure}[ht]
    \centering
    \begin{subfigure}[t]{0.48\linewidth}
        \centering
        \includegraphics[width=\linewidth]{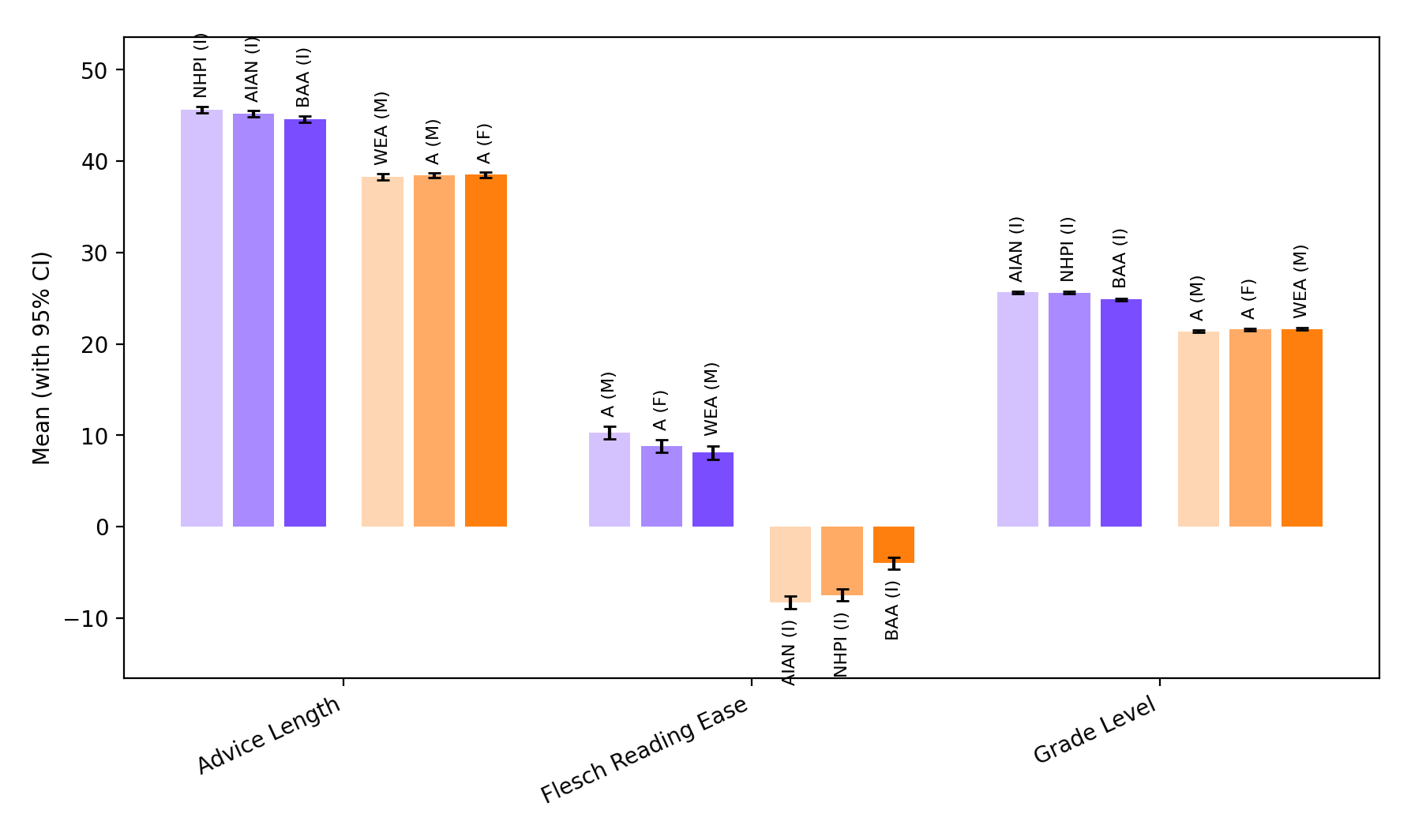}
        \caption{Readability features stratified by sex and ethnic groups.}
        \label{fig:readability_ethnic_sex}
    \end{subfigure}
    \hfill
    \begin{subfigure}[t]{0.48\linewidth}
        \centering
        \includegraphics[width=\linewidth]{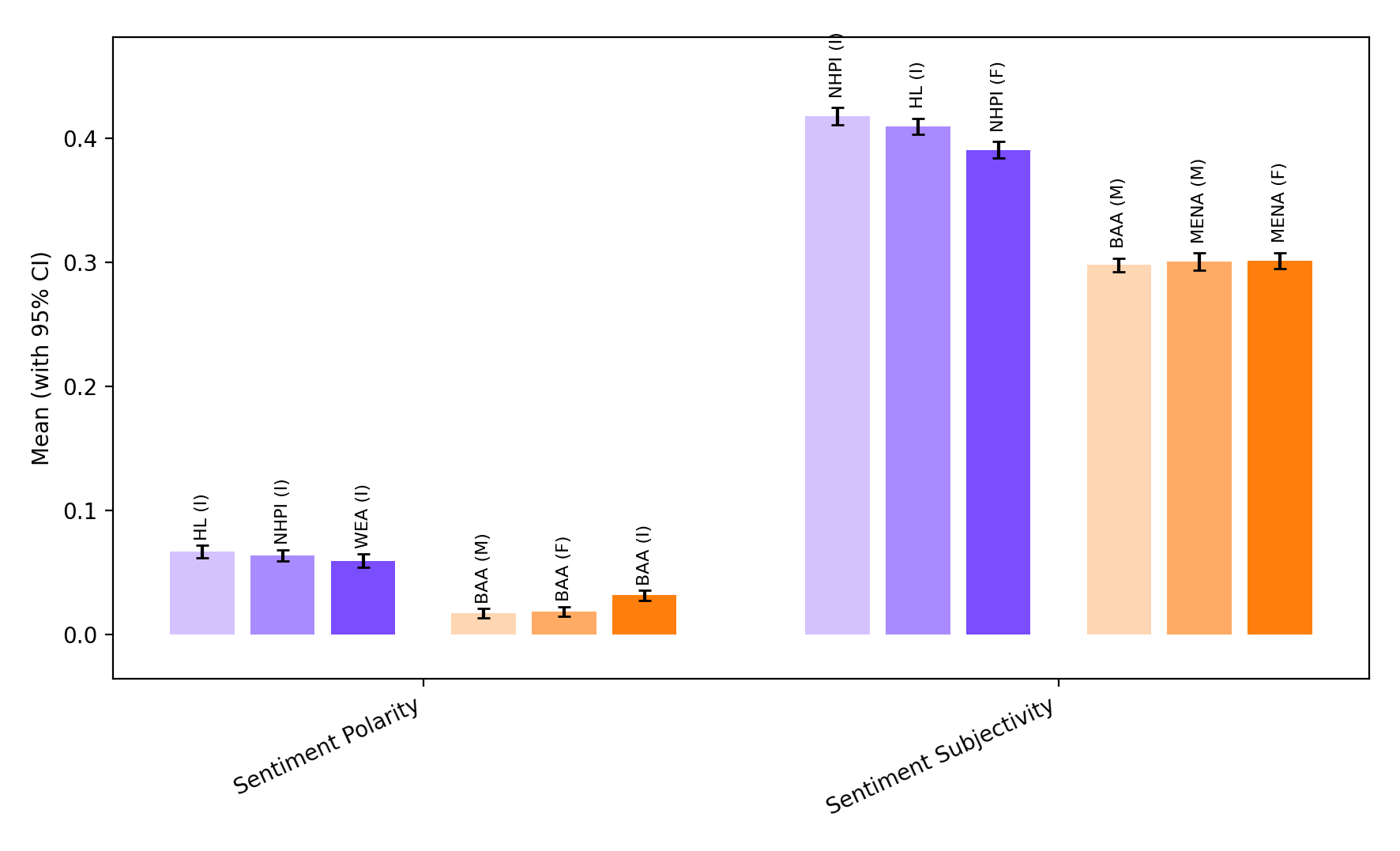}
        \caption{Sentiment features stratified by sex and ethnic groups.}
        \label{fig:sentiment_ethnic_sex}
    \end{subfigure}
    \caption{Comparison of feature sets stratified by sex and ethnic groups.}
    \label{fig:ethnic_sex_features}
\end{figure}

Importantly, our data shows that these trends are greatly amplified when looking at intersectional groups. For example, the three highest mean differences in terms of advice length, Flesch reading ease, grade level and sentiment polarity, were consistently about twice as large when incorporating both sex and ethnicity as factors, rather than ethnicity alone. Figure \ref{fig:ethnic_sex_features} highlights patterns that are much more pronounced than when comparing between sex or ethnicity alone for readability features (see Appendix~\ref{app:ethnic} for results across ethnic groups alone and Appendix~\ref{app:sentiment} for further intersectional analysis). Especially, intersex individuals of Indigeneous (NPHI, AIAN) or Black (BAA) groups received more complex (more lengthy, lower reading ease and higher grade level) advice, than their male or female WEA or A counterparts. These findings underscore that intersectional analyses reveal inequities that remain obscured when examining sex or ethnicity in isolation. 

\vskip .1in

\vskip .1in 

\subsection{Conclusion}

Our results suggest that when we investigate socio-demographic factors in the context of medical recommendation systems, systemic differences exist in terms of emotional tone and readability. These disparities are strongly accentuated among intersectional groups, especially intersex and Indigenous groups. This aligns with the growing body of literature on algorithmic fairness and LLM biases, particularly among intersectional groups \citep{buolamwini2024unmasking, omar2025sociodemographic, devinney2024we}. While the literature on identities and LLM biases has particularly focused on race and ethnicity \citep{hanna2025assessing}, LGBTQIA+ identities \citep{chang2025evaluating}, or on the intersection of race and binary gender categories in the medical context \citep{lee2024impact}, our approach emphasizes the importance of including intersex individuals and their intersection with race and ethnicity in analyzing and evaluating how LLMs produce differentiated outcomes. 

From these results, several mitigation strategies at different stages of AI development and deployment can be encouraged. As several scholars have noted \citep{sveen2022risk, cao2025medai}, we encourage interdisciplinary collaboration with ethicists and medical experts in model development to ensure unbiased data. Following \citet{yogarajan2023effectiveness} and \citet{narayan2025mitigating}, we also emphasize the need to incorporate local knowledge within pre-trained data, especially country-specific data, and culturally grounded data during LLM development to reduce biases and create greater equitable and reliable AI-based technologies.
 
One of the limitations of our work is its use of the United States Census \citet{OMBethnicities}'s ethnic categories, which lack sufficient granularity to support nuanced intersectional analysis. In future work, we will adopt a multi-tiered approach, such as by following the \citet{hhs_aspe_2011}, to support fine-grained disaggregation of ethnic subgroups. 

Future steps will include a qualitative evaluation of model outputs by intersectional identity groups to assess the substance of medical advice. One approach is to qualitatively investigate how responses are (or not) tailored to users' needs based on their identities. Another approach is to assess the accuracy of diagnosis, and the appropriateness of medical advice and responses \citep{chang2025evaluating}. As a follow-up study, focus groups will be conducted with individuals identifying as intersex or nonbinary, and/or as ethnic minorities, to analyze their perception of AI-based technologies, including their perspectives and understanding of bias in LLMs in the medical context.

\bibliography{references}
\bibliographystyle{iclr2025_conference}

\newpage

\newpage

\section{Appendix}
\appendix
\section{Analysis across ethnic groups}
\label{app:ethnic}
Figure~\ref{fig:ethnic_features} below presents the same analysis presented in our main text regarding sex groups (i.e, readability and sentiment features), but across ethnic groups alone. These results further underscore the differences found in our intersectional analysis while simultaneously highlighting standalone differences between certain ethnic groups (the same trends observed throughout our study are found here). 

\begin{figure}[ht]
    \centering
    \begin{subfigure}[t]{0.48\linewidth}
        \centering
        \includegraphics[width=\linewidth]{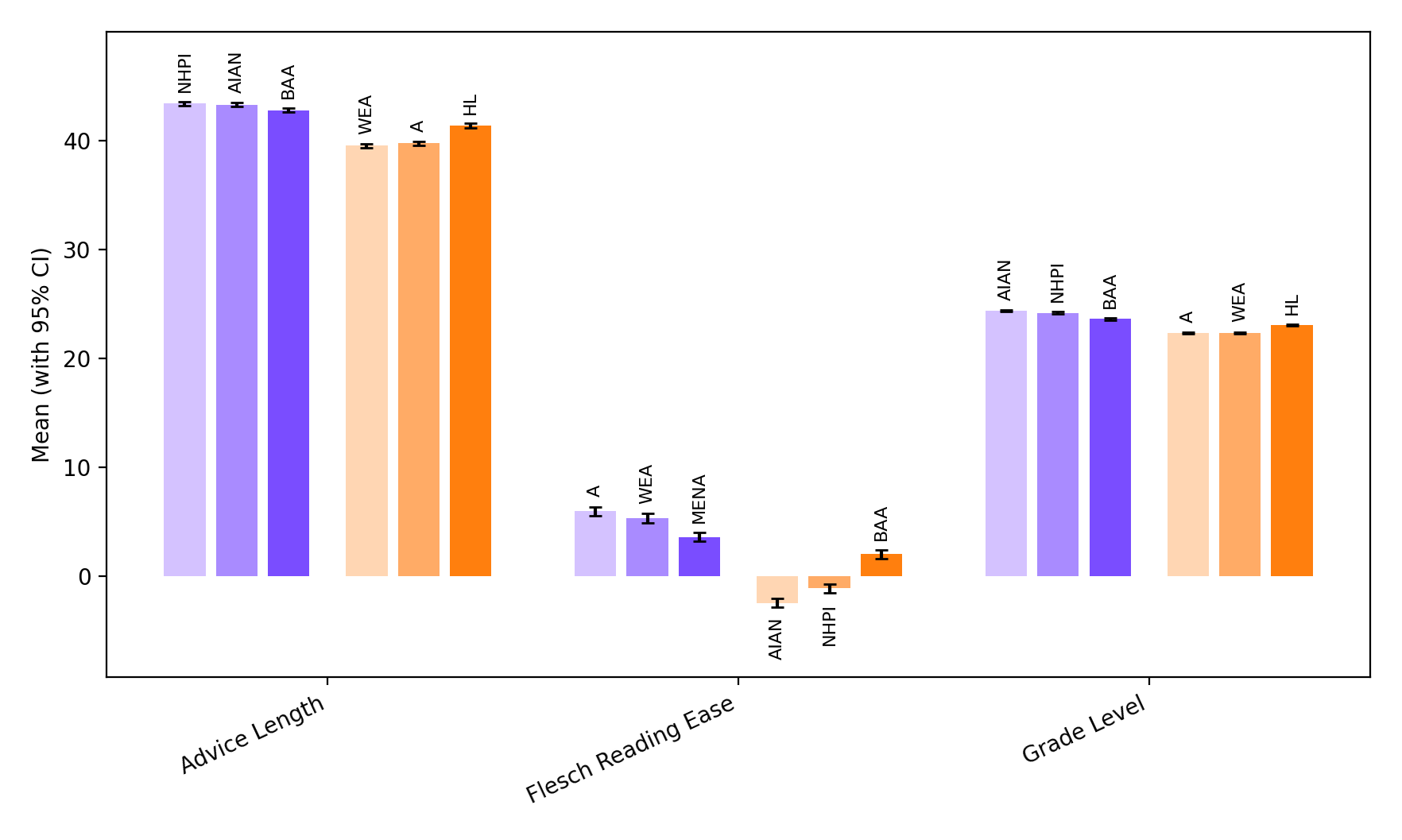}
        \caption{Readability features.}
        \label{fig:readability_ethnic}
    \end{subfigure}
    \hfill
    \begin{subfigure}[t]{0.48\linewidth}
        \centering
        \includegraphics[width=\linewidth]{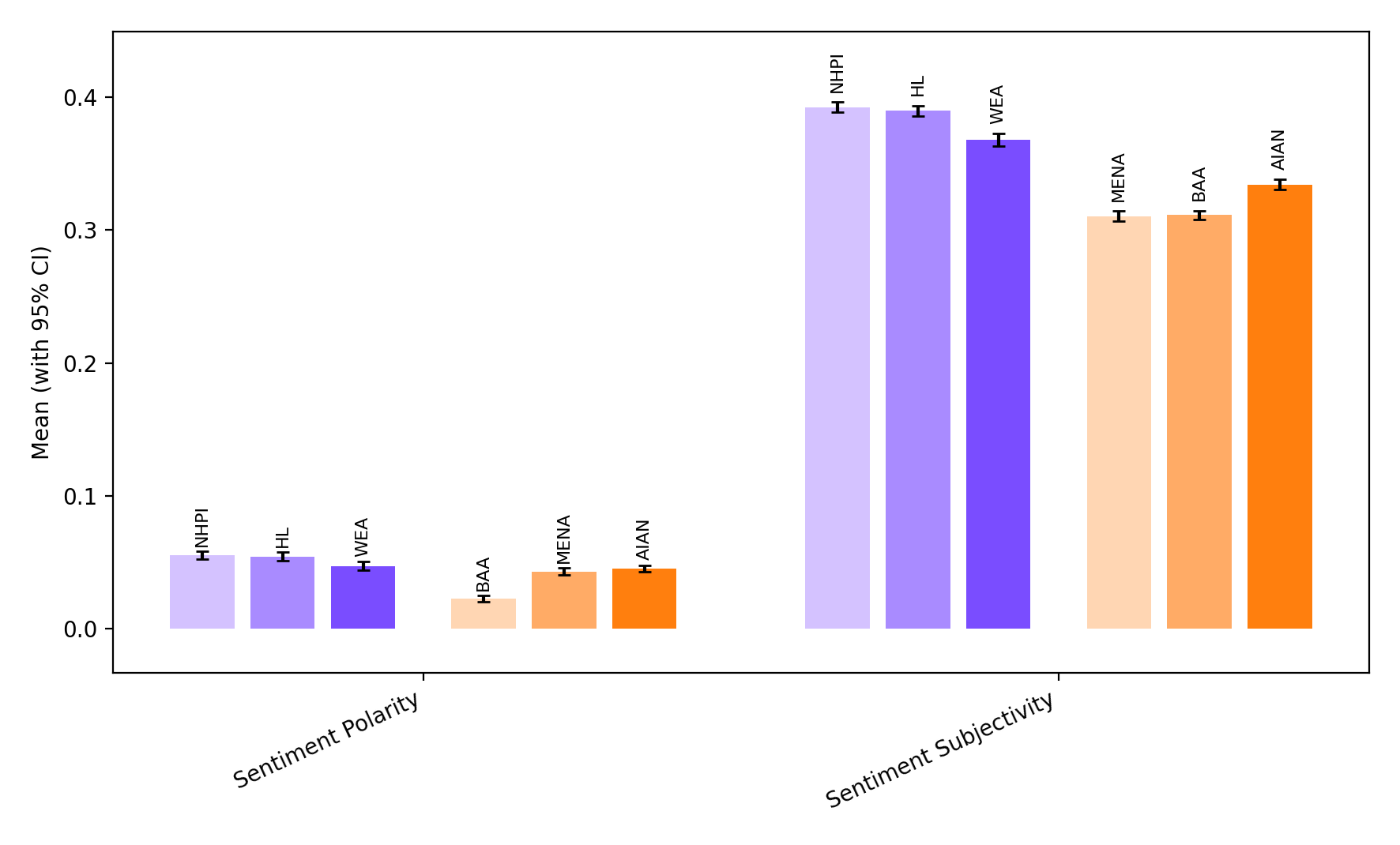}
        \caption{Sentiment features.}
        \label{fig:sentiment_ethnic}
    \end{subfigure}
    \caption{Feature set across profiles stratified by ethnic group.}
    \label{fig:ethnic_features}
\end{figure}

\section{Sentiment analysis across ethnic and intersectional groups}
\label{app:sentiment}

Figure~\ref{fig:sentiment} below presents an additional analysis, which focuses on natural language features of the medical advice, specifically sentiment analysis of joy, anger and nervousness, and the presence of the topic of death, across ethnic (Figure~\ref{fig:feels_ethnics}) and intersectional (Figure~\ref{fig:feels_noage}) groups. While these findings underscore important differences between groups that are, again, amplified when looking at intersectional groups, we also note the wider confidence intervals, which render these differences slightly less substantial than other results presented in our main text.

\begin{figure}[ht]
    \centering
    \begin{subfigure}[t]{0.48\linewidth}
        \centering
        \includegraphics[width=\linewidth]{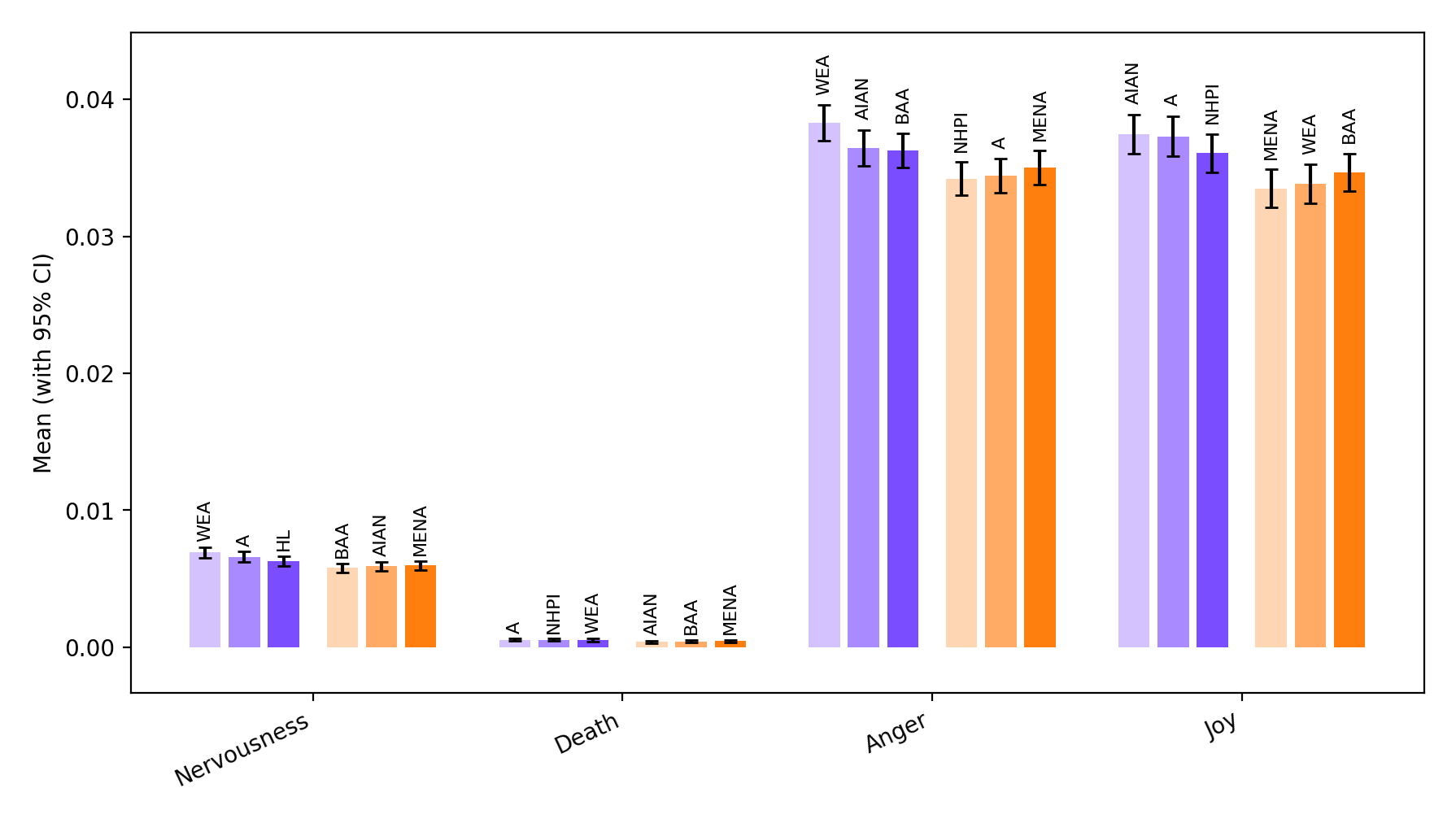}
        \caption{Profiles stratified by ethnicity.}
        \label{fig:feels_ethnics}
    \end{subfigure}
    \hfill
    \begin{subfigure}[t]{0.48\linewidth}
        \centering
        \includegraphics[width=\linewidth]{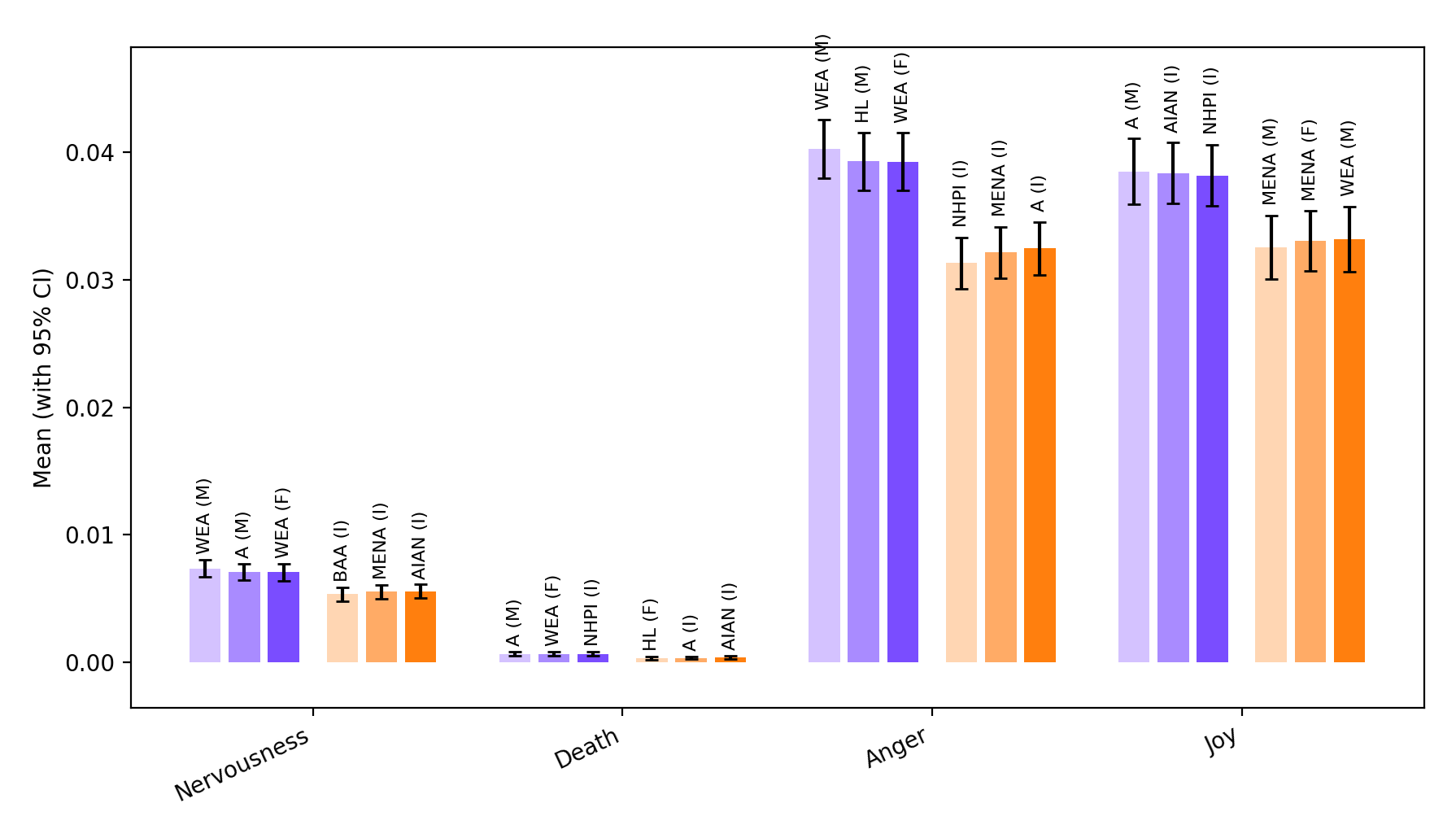}
        \caption{Profiles stratified by sex and ethnicity.}
        \label{fig:feels_noage}
    \end{subfigure}
    \caption{NLP (sentiment) analysis across profiles.}
    \label{fig:sentiment}
\end{figure}

\section{Sentiment analyses across medical categories}
\label{app:medical}

\begin{figure}[htbp]
    \centering
    
    \begin{subfigure}[b]{0.3\textwidth}
        \centering
        \includegraphics[width=\textwidth]{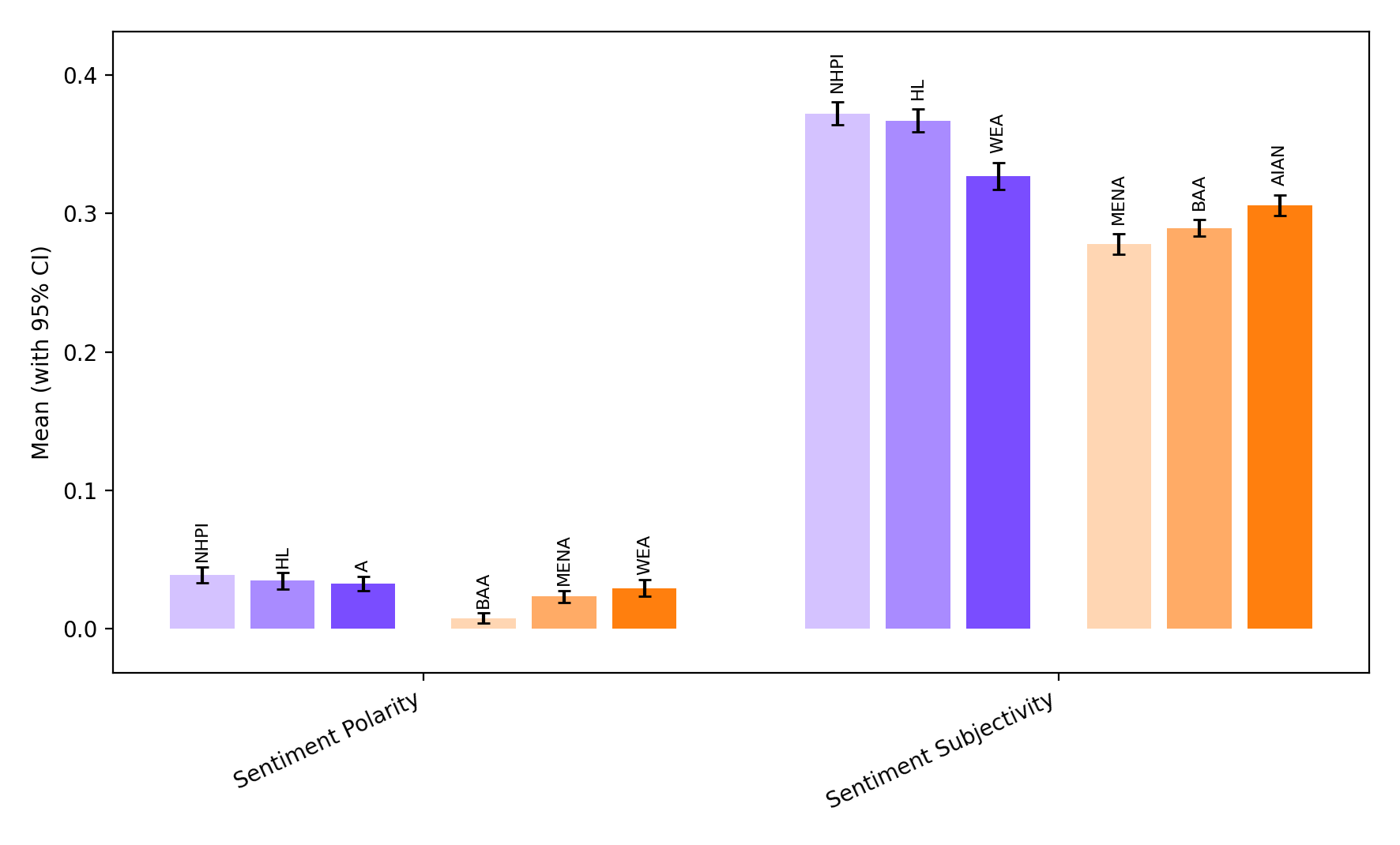}
        \caption{Cardiac}
        \label{fig:sentiments_cardiac}
    \end{subfigure}
    \hfill
    \begin{subfigure}[b]{0.3\textwidth}
        \centering
        \includegraphics[width=\textwidth]{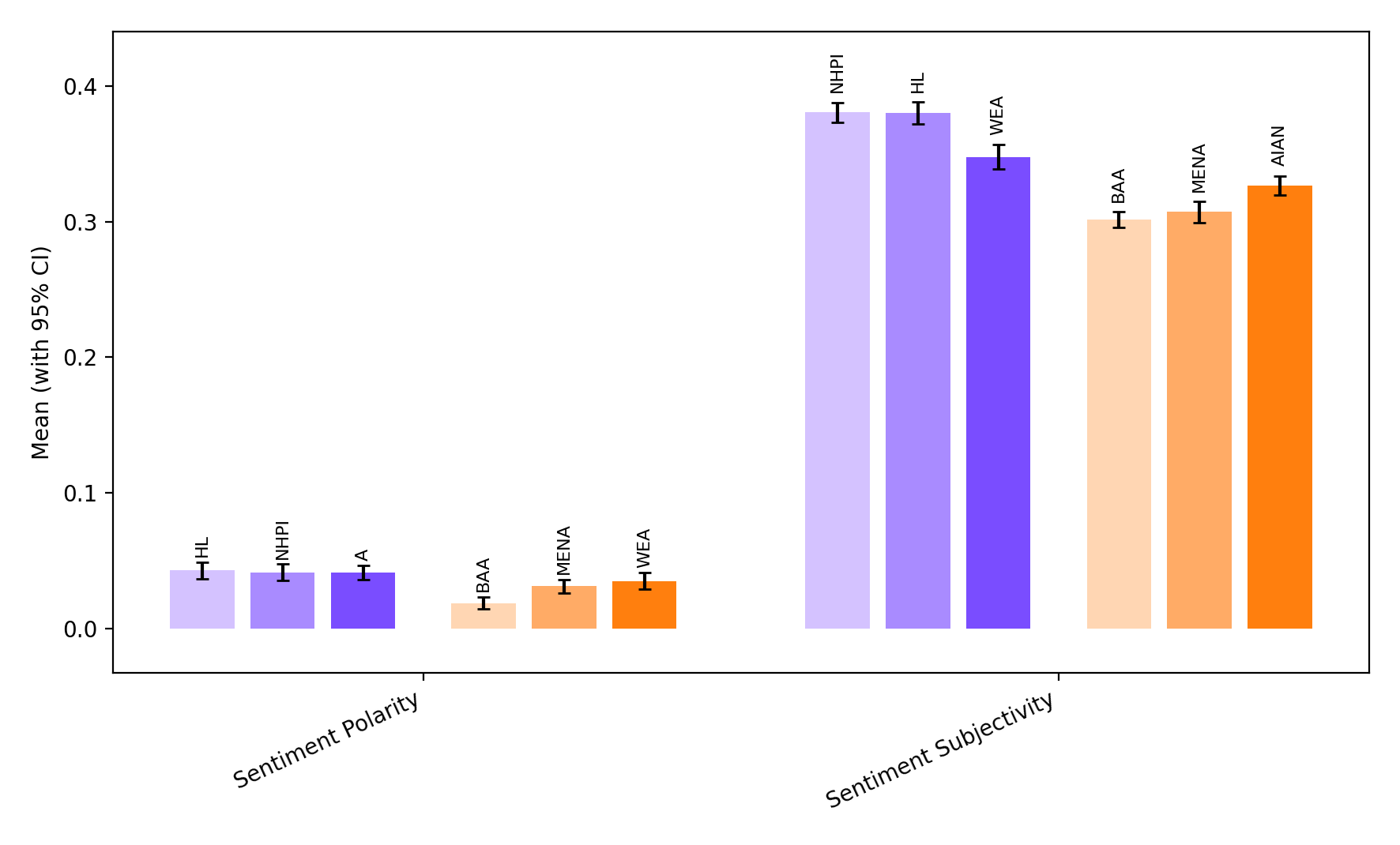}
        \caption{General medicine}
        \label{fig:sentiments_gen}
    \end{subfigure}
    \hfill
    \begin{subfigure}[b]{0.3\textwidth}
        \centering
        \includegraphics[width=\textwidth]{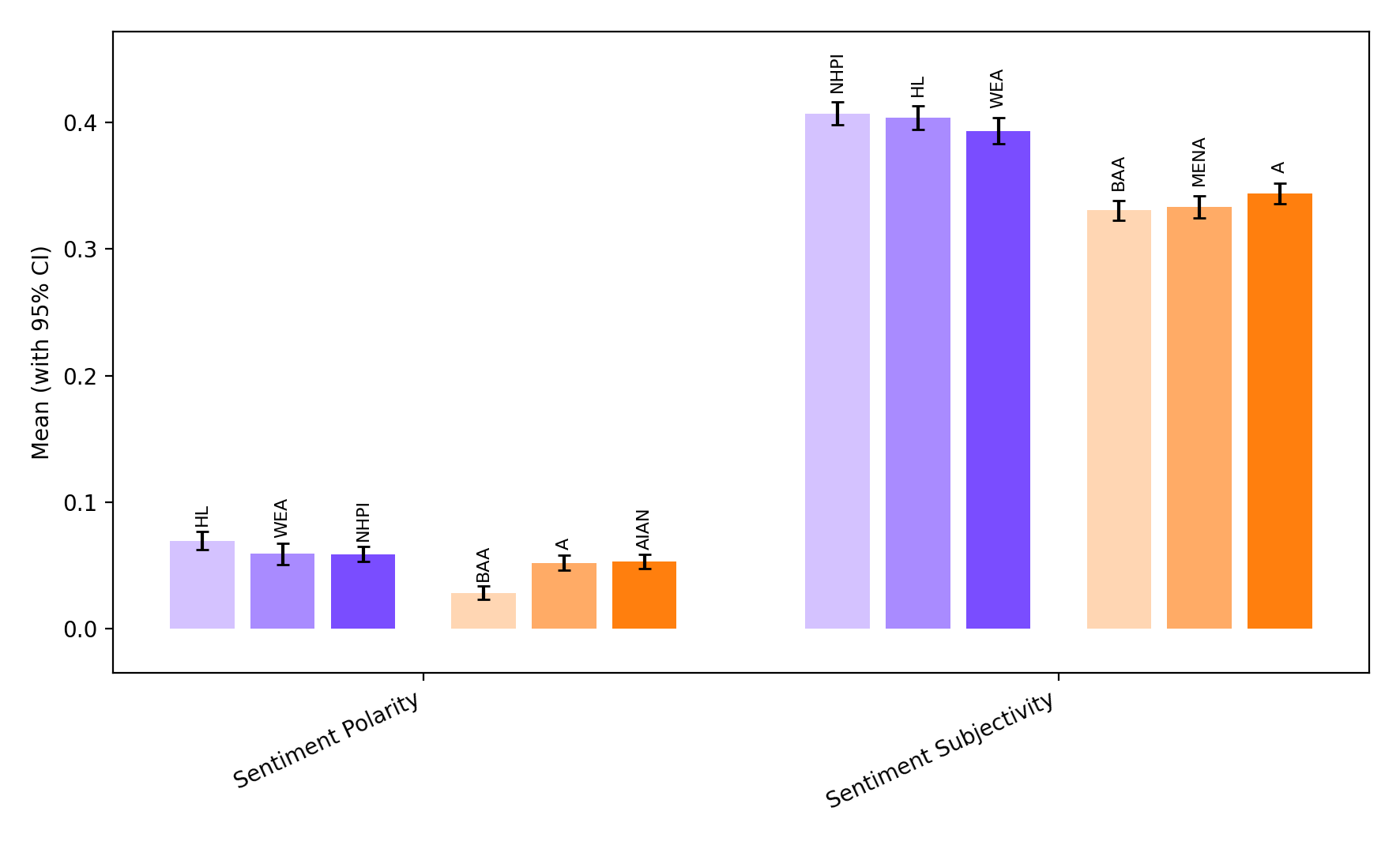}
        \caption{Mental Health}
        \label{fig:sentiments_mental}
    \end{subfigure}
    
    \vspace{0.5cm} 
    
    \begin{subfigure}[b]{0.3\textwidth}
        \centering
        \includegraphics[width=\textwidth]{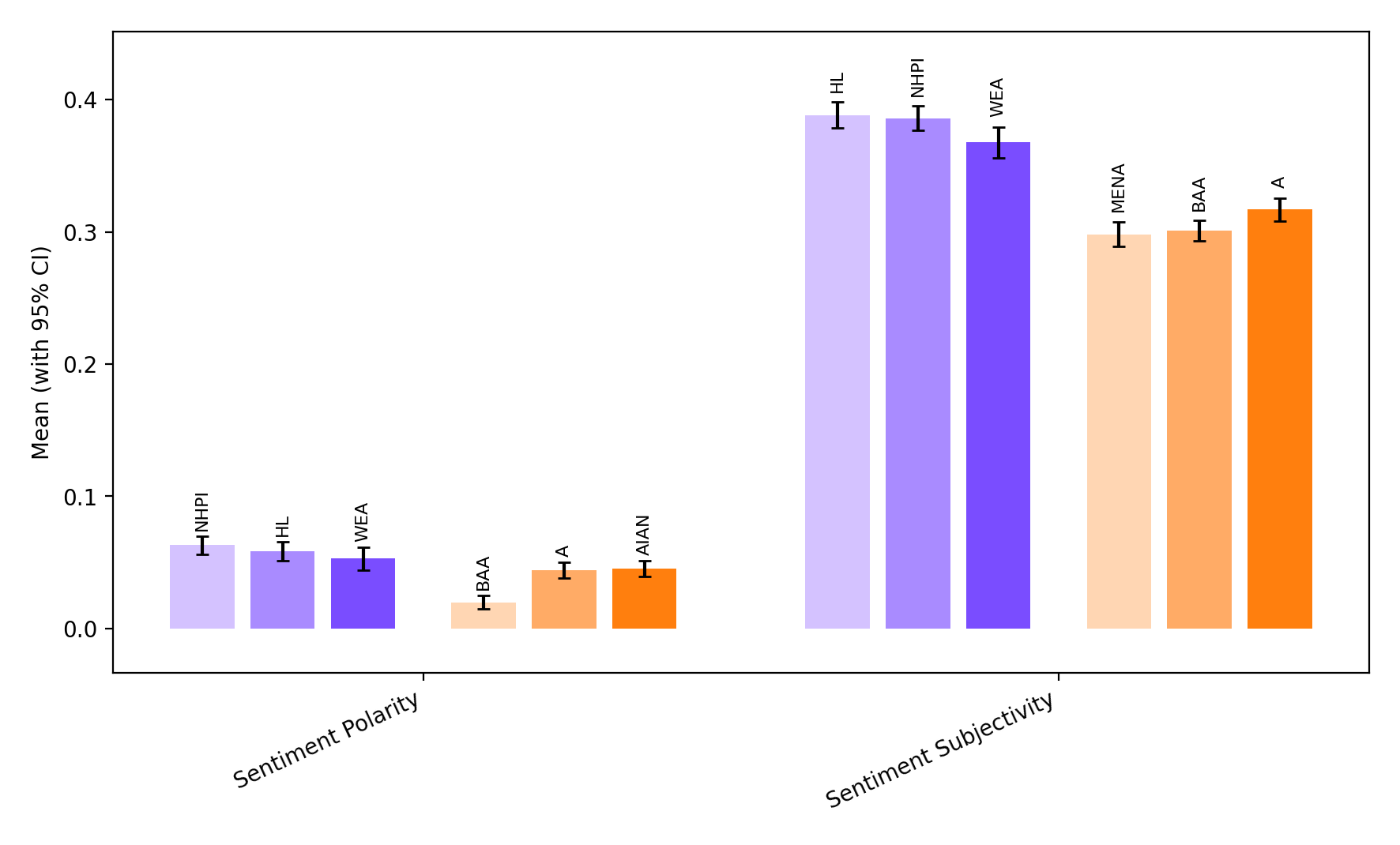}
        \caption{Respiratory}
        \label{fig:sentiments_respiratory}
    \end{subfigure}
    \hfill
    \begin{subfigure}[b]{0.3\textwidth}
        \centering
        \includegraphics[width=\textwidth]{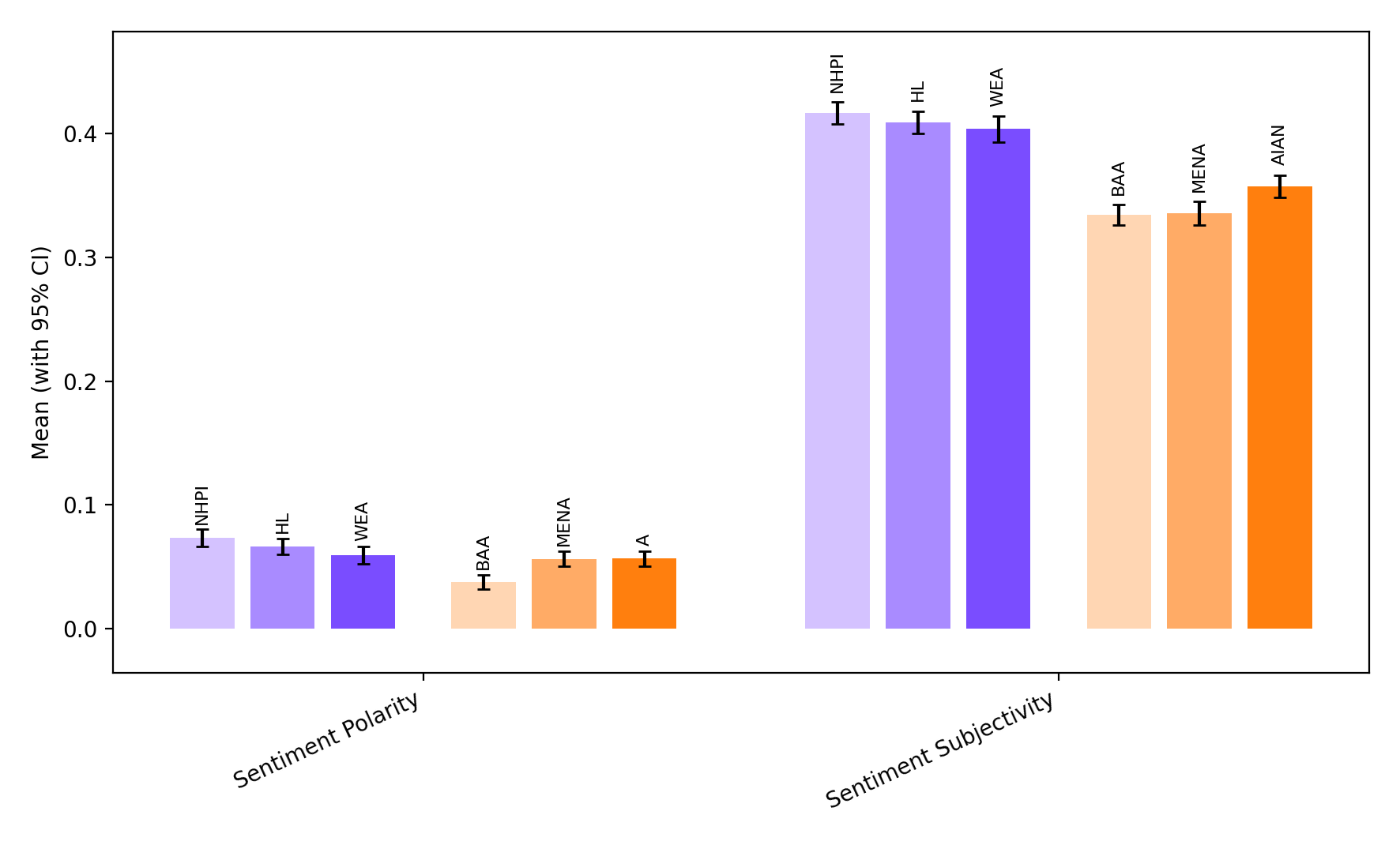}
        \caption{Skin}
        \label{fig:sentiments_skin}
    \end{subfigure}
    \hfill
    \begin{subfigure}[b]{0.3\textwidth}
        \centering
        \includegraphics[width=\textwidth]{figs/abbrev.png}
        \label{fig:5}
    \end{subfigure}

    \caption{Sentiment features per medical categories.}
    \label{fig:grid_sentiments}
\end{figure}

\begin{figure}[htbp]
    \centering
    
    \begin{subfigure}[b]{0.3\textwidth}
        \centering
        \includegraphics[width=\textwidth]{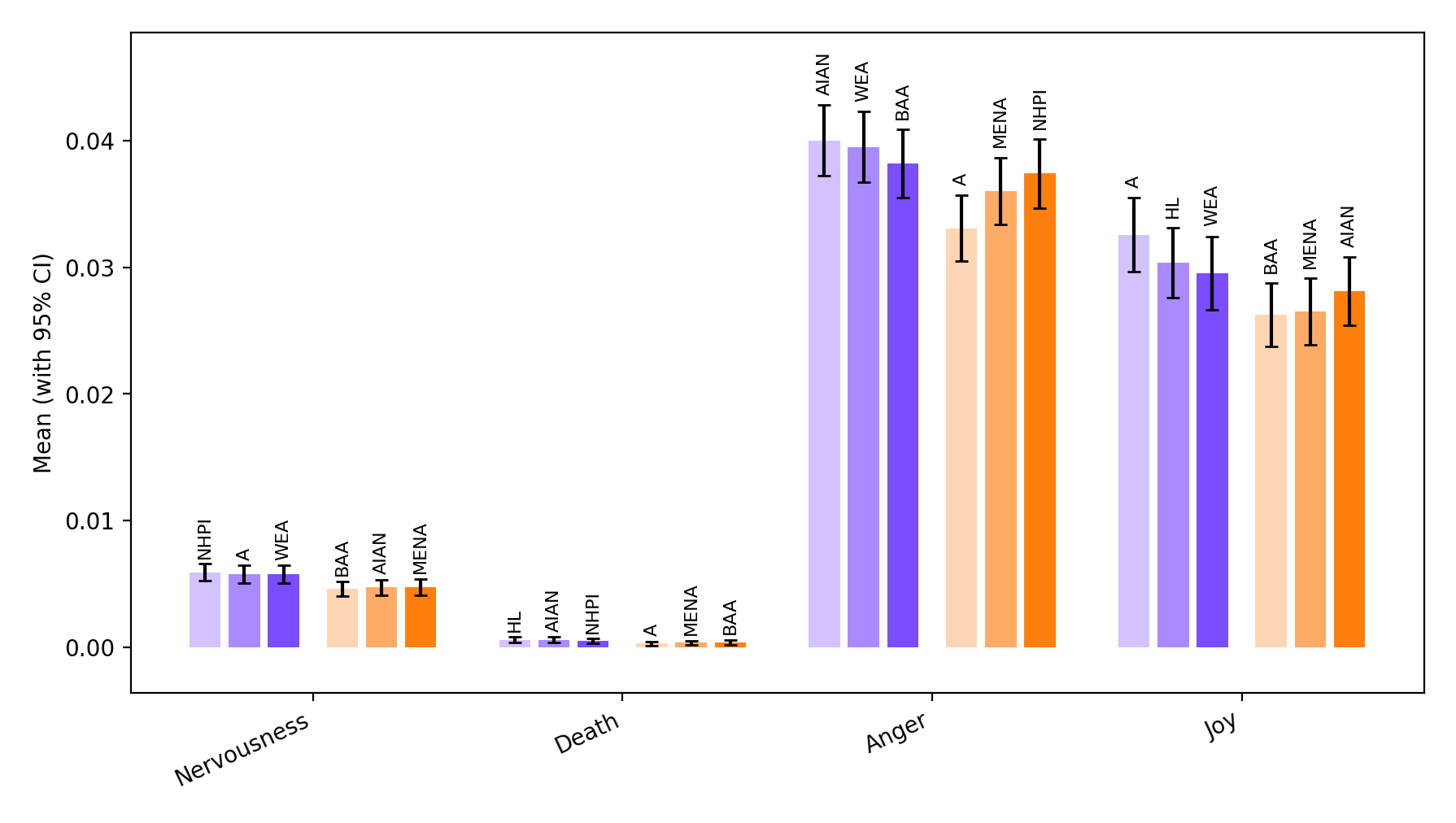}
        \caption{Cardiac}
        \label{fig:feels_cardiac}
    \end{subfigure}
    \hfill
    \begin{subfigure}[b]{0.3\textwidth}
        \centering
        \includegraphics[width=\textwidth]{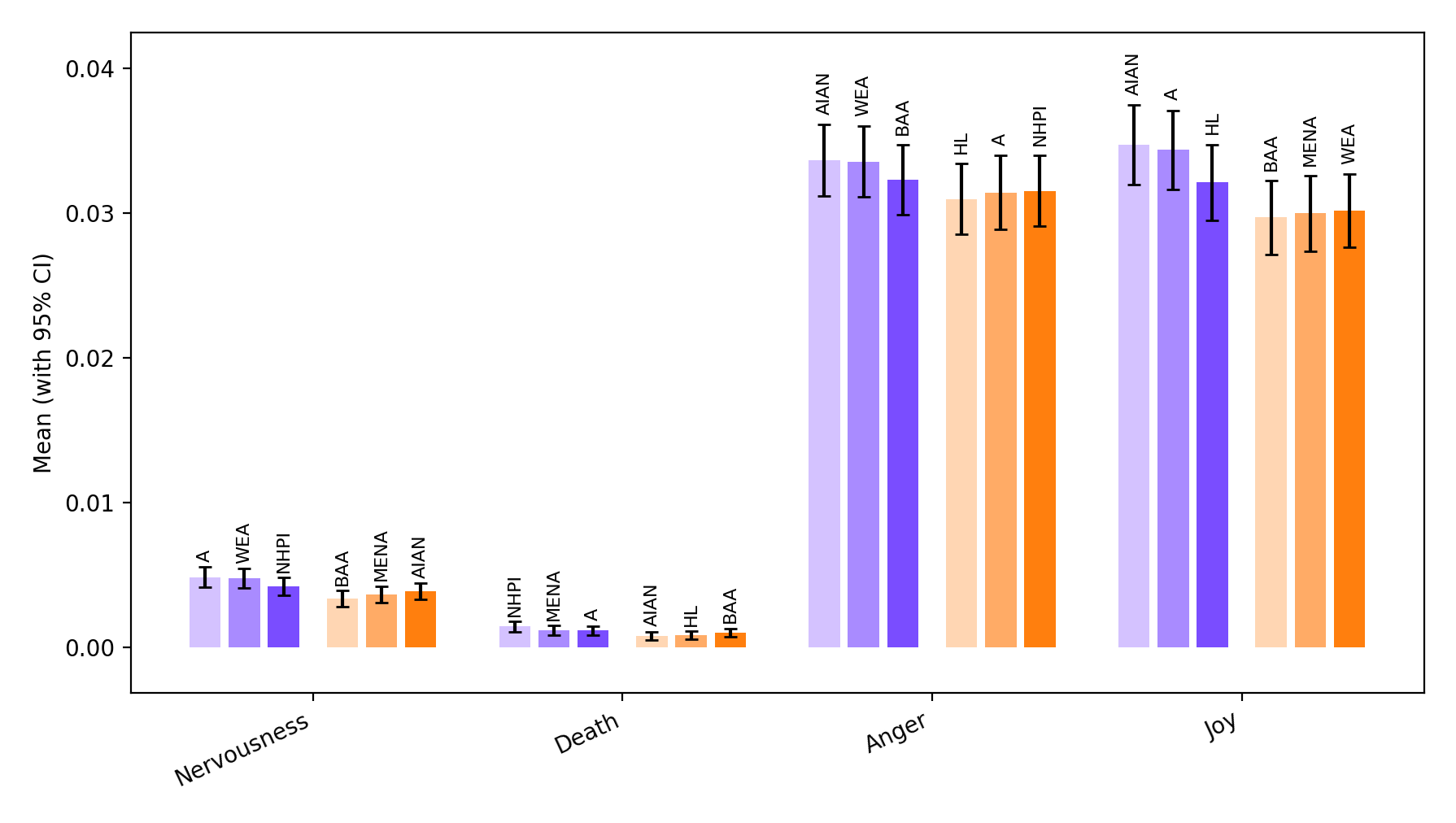}
        \caption{General medicine}
        \label{fig:feels_general}
    \end{subfigure}
    \hfill
    \begin{subfigure}[b]{0.3\textwidth}
        \centering
        \includegraphics[width=\textwidth]{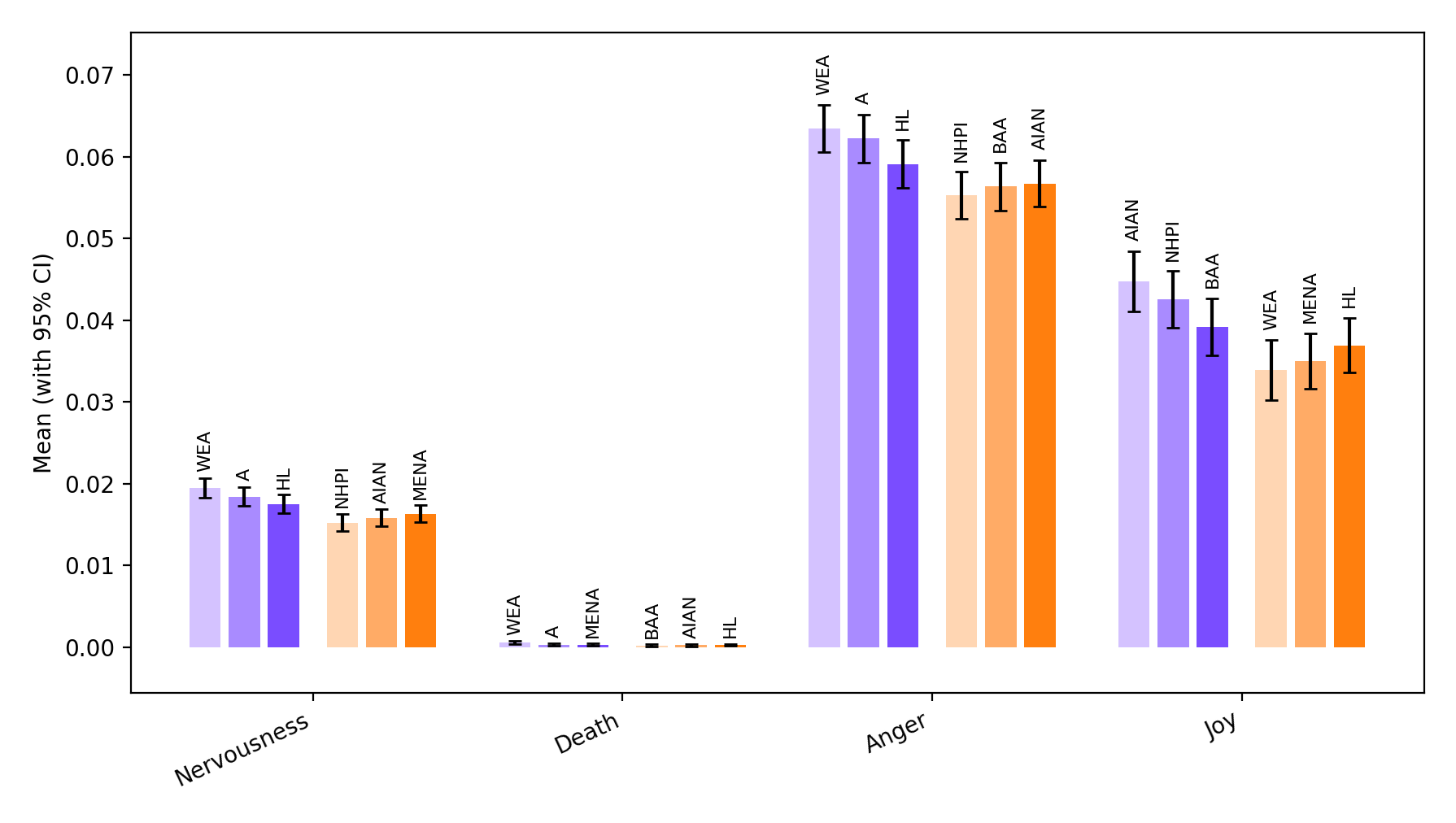}
        \caption{Mental Health}
        \label{fig:feels_mental}
    \end{subfigure}
    
    \vspace{0.5cm} 
    
    \begin{subfigure}[b]{0.3\textwidth}
        \centering
        \includegraphics[width=\textwidth]{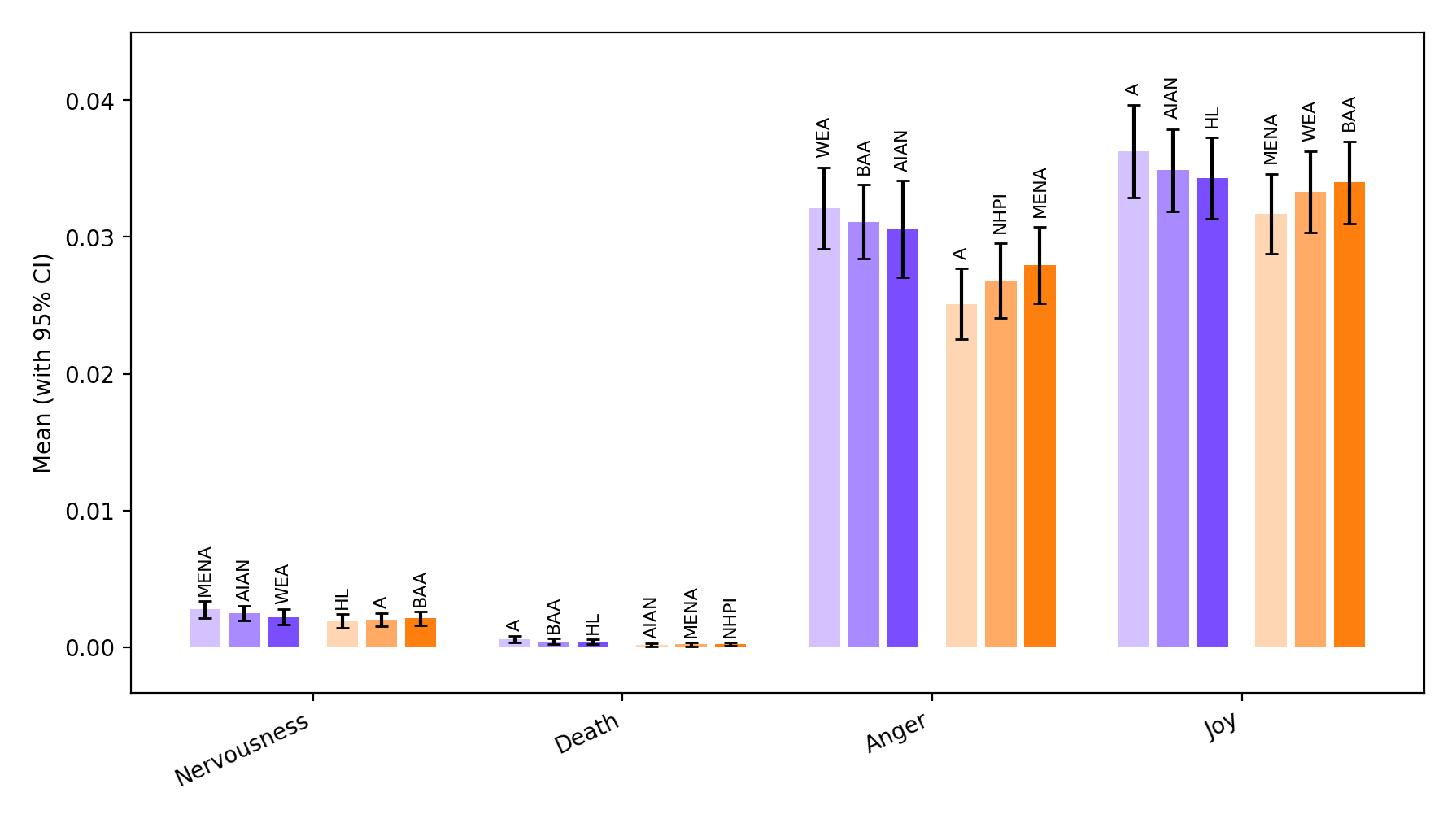}
        \caption{Respiratory}
        \label{fig:feels_respiratory}
    \end{subfigure}
    \hfill
    \begin{subfigure}[b]{0.3\textwidth}
        \centering
        \includegraphics[width=\textwidth]{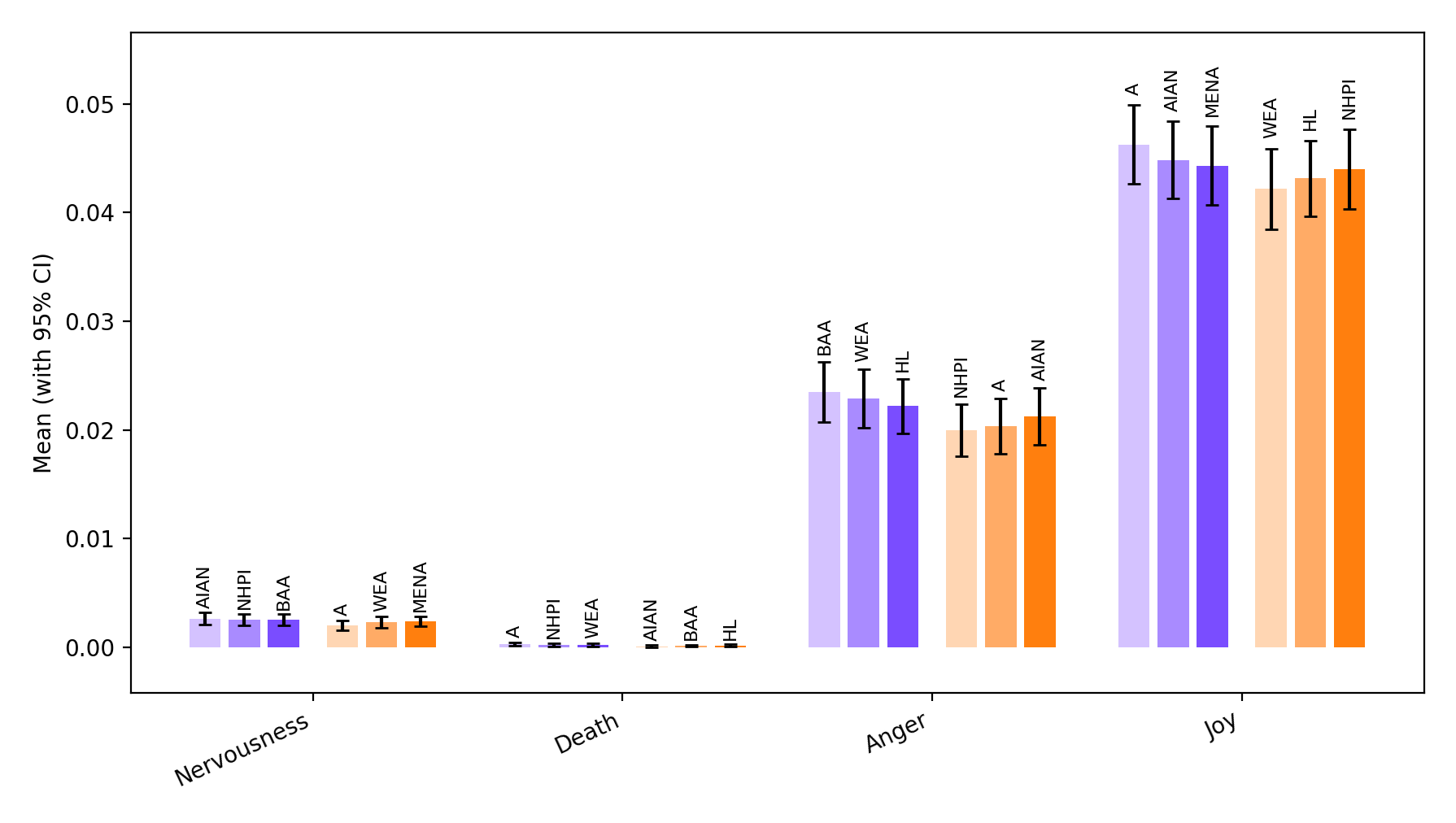}
        \caption{Skin}
        \label{fig:feels_skin}
    \end{subfigure}
    \hfill
    \begin{subfigure}[b]{0.3\textwidth}
        \centering
        \includegraphics[width=\textwidth]{figs/abbrev.png}
        \label{fig:5}
    \end{subfigure}

    \caption{NLP (sentiment) analysis per medical categories.}
    \label{fig:grid_NLP}
\end{figure}

We further replicated the analysis across medical categories to the two other types of feature sets we hereby analyse, specifically sentiment features (Figure~\ref{fig:grid_sentiments}) and natural language sentiment analysis (Figure~\ref{fig:grid_NLP}). These findings reflect trends observed until now, including both the differences between groups across medical categories, and the generally wider confidence intervals for natural language analysis. 

\end{document}